\documentclass{article}

\usepackage[utf8]{inputenc}
\usepackage[T1]{fontenc}
\usepackage{lmodern}
\usepackage{authblk}
\usepackage{setspace}
\usepackage[margin=1.25in]{geometry}
\usepackage{graphicx}
\graphicspath{{figure/}}
\usepackage{subcaption}
\usepackage{microtype}
\usepackage{amsmath}
\usepackage{amssymb}
\usepackage{amsfonts}
\usepackage{bm}
\usepackage{booktabs}
\usepackage{multirow}
\usepackage{array}
\usepackage{tabularx}
\usepackage{makecell}
\usepackage{ragged2e}
\usepackage{siunitx}
\usepackage{enumitem}
\usepackage[hidelinks]{hyperref}

\usepackage[
  style=nejm,
  citestyle=numeric-comp,
  sorting=none
]{biblatex}
\newcommand{\safeincludegraphics}[2][]{%
  \IfFileExists{#2}{\includegraphics[#1]{#2}}{%
    \fbox{\parbox[c][0.24\textheight][c]{0.9\linewidth}{\centering\small Missing figure file: \texttt{\detokenize{#2}}}}%
  }%
}
\newcolumntype{L}[1]{>{\RaggedRight\arraybackslash}p{#1}}
\newcolumntype{C}[1]{>{\Centering\arraybackslash}p{#1}}
\newcolumntype{Y}{>{\RaggedRight\arraybackslash}X}
\title{BioMamba: Domain-Adaptive Biomedical Language Models}

\author[1,*]{Ling Yue}
\author[1,*]{Mingzhi Zhu}
\author[1]{Sixue Xing}
\author[2,$\dag$]{Yunning Cao}
\author[3,$\dag$]{Yanbo Wang}
\author[3]{Shimin Shan}
\author[4]{Jinfei Liu}
\author[5]{Vijil Chenthamarakshan}
\author[1]{Shaowu Pan}
\author[5,$\dag$]{Payel Das}
\author[6]{Tianfan Fu}

\affil[1]{Rensselaer Polytechnic Institute, Troy, USA.}
\affil[2]{Jiaxing New Jies Thermal Power Co. Ltd., Jiaxing, China.}
\affil[3]{North University of China, Taiyuan, China.}
\affil[4]{Zhejiang University, Hangzhou, China.}
\affil[5]{IBM Research, Yorktown Heights, USA.}
\affil[6]{Nanjing University, Nanjing, China.}
\affil[$\dag$]{Address correspondence to: \href{mailto:caoyn@zjmi.com}{caoyn@zjmi.com} (Y.C.); \href{mailto:yanbowang@nuc.edu.cn}{yanbowang@nuc.edu.cn} (Y.W.); \href{mailto:daspa@us.ibm.com}{daspa@us.ibm.com} (P.D.);}
\affil[*]{These authors contributed equally to this work.}

\date{}

\begin{document}

\maketitle

\begin{abstract}
\noindent\textbf{Background.} Biomedical language models should improve performance on biomedical text while retaining general-language-modeling fluency. For Mamba-based models, this trade-off has not been systematically studied across biomedical literature and clinical text.

\noindent\textbf{Methods.} We developed BioMamba, a family of biomedical Mamba2 models at five scales (130M, 370M, 780M, 1.3B, and 2.7B) obtained by continued pretraining of released public Mamba2 checkpoints on a balanced 80\%/10\%/10\% mixture of PubMed abstracts, the Colossal Clean Crawled Corpus (C4), and Wikipedia. The architecture is unchanged from Mamba2; the contribution is the adaptation recipe and the accompanying open-weight checkpoints. We evaluated internal language modeling on fixed held-out sets, out-of-domain multiple-choice benchmarks, and three downstream tasks across multiple model scales: clinical note completion and discharge summary generation on MIMIC-IV-Note, and biomedical yes/no question answering on BioASQ and PubMedQA.

\noindent\textbf{Results.} Across five scales, BioMamba consistently lowered PubMed perplexity, improved Wikipedia-style held-out perplexity by 1.46--4.72 PPL, and left C4 perplexity essentially unchanged ($|\Delta\mathrm{PPL}| \le 0.17$). On six out-of-domain multiple-choice benchmarks, BioMamba stayed within $\pm 3$ percentage points of Mamba2 with no systematic regression. After supervised fine-tuning, BioMamba+SFT matched or exceeded Mamba2+SFT on MIMIC-IV note completion and discharge summary generation at every evaluated scale (paired-bootstrap $P(\Delta > 0) \ge 0.997$ at 2.7B), and improved PubMedQA at every scale. The strongest model (BioMamba-2.7B) reached a PubMed perplexity of 5.28 and accuracies of 90.24\% and 73.00\% on BioASQ and PubMedQA, respectively.

\noindent\textbf{Conclusions.} A balanced domain-adaptive continued pretraining recipe strengthens Mamba2 language models on biomedical literature and clinical text while preserving general-language-modeling fluency.
\end{abstract}

\section{Introduction}

Biomedical research and health data science increasingly rely on unstructured text, including PubMed abstracts, full-text articles, clinical guidelines, trial reports, and clinical documentation. Domain-adapted language models are therefore important for biomedical information extraction, evidence synthesis, summarization, completion, and question answering \cite{Jin2021BiomedicalQA}. Transformer-based biomedical models such as BioBERT \cite{lee2020biobert}, SciBERT \cite{beltagy2019scibert}, PubMedBERT \cite{gu2021domain}, BioGPT \cite{luo2022biogpt}, SciFive \cite{Phan2021SciFiveAT}, and BioBART \cite{Yuan2022BioBARTPA} have shown that adapting general-purpose language models to biomedical corpora can yield strong performance. In clinical natural language processing, ClinicalBERT \cite{Alsentzer2019PubliclyAC}, GatorTron \cite{Yang2022GatorTronAL}, and ClinicalT5 \cite{Lu2022ClinicalT5AG} further demonstrate the value of domain adaptation for clinical text. Recent work in health data science and public-health artificial intelligence has also emphasized that models used in biomedical workflows should be accurate, transparent, and practically useful \cite{lu2024uncertainty,li2024federated,tao2025litautoscreener,zhang2023emerging,llm_healthcare_review_2025,ai_public_health_2025}.

Most biomedical language models still use Transformer self-attention, which has quadratic cost in sequence length \cite{vaswani2017attention}. This cost becomes restrictive when biomedical inputs are long, terminology-dense, and distributed across multiple sentences or sections. State space models, especially structured state space models \cite{Gu2021EfficientlyML}, offer an alternative. Mamba introduced selective state space layers for linear-time sequence modeling, and later work further strengthened the connection between state space models and attention while preserving favorable scaling for long contexts \cite{gu2023mamba,dao2024transformers}.

Despite this progress, biomedical adaptation of Mamba models remains limited. Biomedical language modeling is still dominated by Transformer families and related open medical language models \cite{lee2020biobert,gu2021domain,luo2022biogpt,bolton2024biomedlm,Wu2023PMCLLaMATB,Labrak2024BioMistralAC,Chen2023MEDITRON70BSM}, and the few reported Mamba-based biomedical adaptations have focused on narrower settings such as clinical notes \cite{yang2024clinicalmamba}. Second, domain adaptation risks catastrophic forgetting: a model may improve on biomedical text while losing general-language-modeling fluency, which is a particular concern since biomedical writing often mixes specialized terminology with broader scientific language. This trade-off has been documented in domain-adaptive pretraining literature~\cite{Gururangan2020DontSP,Parmar2024ReuseDR}, and in work explicitly targeting general knowledge preservation~\cite{Ke2023AdaptingAL}.

We address this gap with BioMamba, a family of biomedical models built by continued pretraining of released Mamba2 checkpoints. Our contribution is a reproducible biomedical continued pretraining recipe together with a family of open-weight BioMamba checkpoints at 130M, 370M, 780M, 1.3B, and 2.7B parameters; the architecture itself is unchanged from Mamba2. The main focus is therefore the adaptation strategy, its empirical effects, and the accompanying resource release rather than an architectural modification. This study makes three contributions:
\begin{itemize}
\item A public release of the training and evaluation code and all five checkpoints at \url{https://github.com/lingyue404/BioMamba} under the Apache-2.0 license, so that the biomedical natural language processing community can directly use or extend the models.
\item A systematic investigation of the continued pretraining pipeline showing that an appropriate 80\%/10\%/10\% PubMed/C4/Wikipedia mixture can enhance biomedical language modeling while preserving general-language-modeling fluency and mitigating catastrophic forgetting.
\item Controlled base-versus-adapted experiments showing that BioMamba achieves consistent improvements across downstream tasks, including biomedical question answering (PubMedQA, BioASQ), clinical note completion, and discharge summary generation on MIMIC-IV.
\end{itemize}

\section{Methods}
\label{sec:methods}

\subsection{Overview}

\begin{figure}[t]
\centering
\safeincludegraphics[width=\linewidth]{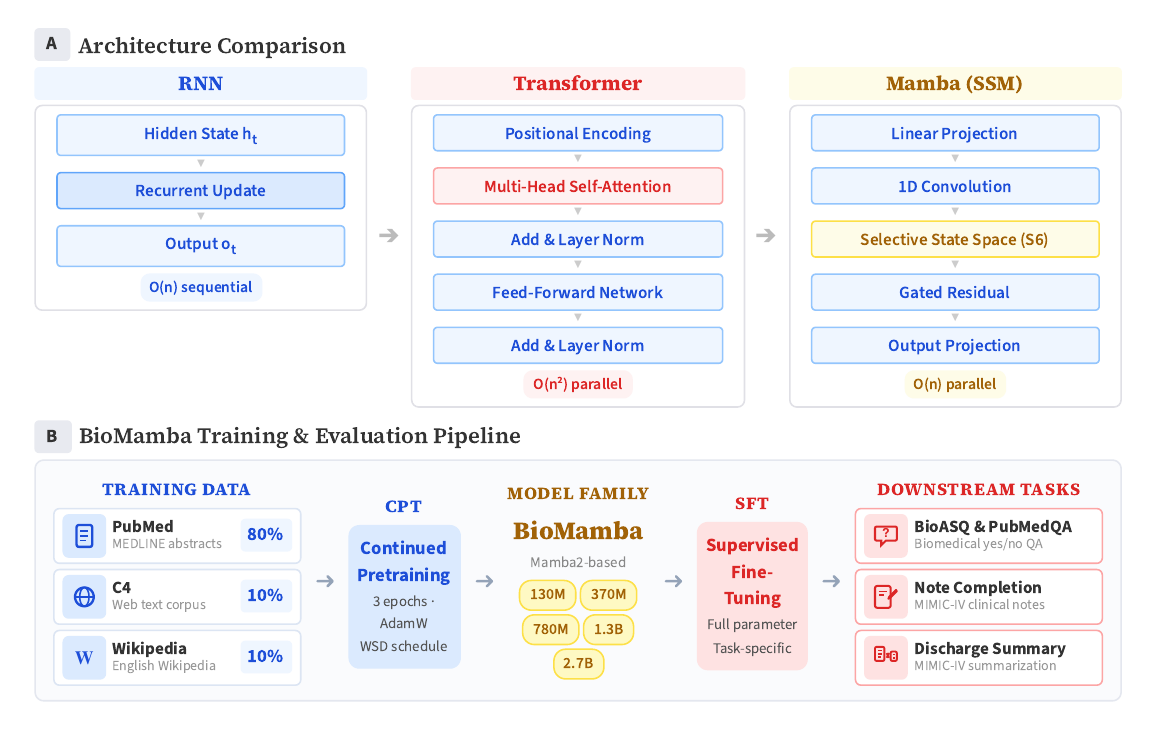}
\caption{Overview of BioMamba. (A)~Architecture comparison: RNN, Transformer, and Mamba (SSM), with key distinguishing components highlighted. (B)~Training and evaluation pipeline: continued pretraining on a mixed PubMed/C4/Wikipedia corpus produces BioMamba models at five scales (130M--2.7B), which are then fine-tuned for biomedical question answering, clinical note completion, and discharge summary generation. WSD: Warmup--Stable--Decay learning-rate schedule.}
\label{fig:biomamba}
\end{figure}

This study tests whether biomedical continued pretraining \cite{Gururangan2020DontSP,Parmar2024ReuseDR} can improve Mamba language models while retaining general-language-modeling fluency. Figure~\ref{fig:biomamba} shows the overall workflow. We started from public Mamba2 checkpoints, continued pretraining on a mixed corpus of PubMed, C4, and Wikipedia, evaluated language-model quality on fixed in-domain and out-of-domain validation sets, and then fine-tuned the resulting models on clinical note completion, discharge summary generation, and biomedical question answering.

Before analysis, we specified which comparisons would be treated as primary. The primary comparison is between each base model and its BioMamba counterpart under a shared tokenizer and fixed evaluation sets. These comparisons isolate the effect of biomedical adaptation most directly. Comparisons with public checkpoints are included only for context because tokenizers and prior training histories differ across released models.

\subsection{Model family and training data}

BioMamba was obtained by continued pretraining of released Hugging Face Mamba2 checkpoints from \texttt{state-spaces/mamba2-*} \cite{gu2023mamba,dao2024transformers} at five model scales spanning 130M--2.7B parameters. To keep tokenization fixed across internal comparisons, all models use the shared GPT-NeoX tokenizer \cite{black2022gpt} from \texttt{state-spaces/mamba-2.8b-hf}, which has a vocabulary of 50,280 tokens.

All BioMamba variants were trained autoregressively with the standard next-token objective,
\begin{equation}
\mathcal{L}_{\mathrm{LM}} = - \sum_{t=1}^{T} \log P(x_t \mid x_{<t}; \theta),
\end{equation}
where $x_t$ is the token at position $t$, $x_{<t}$ denotes the preceding context, and $\theta$ represents model parameters.

PubMed abstracts indexed in Medical Literature Analysis and Retrieval System Online (MEDLINE) provided the biomedical data \cite{canese2013pubmed}, and we added C4 \cite{Dodge2021DocumentingLW} and English Wikipedia \cite{Merity2016PointerSM} in smaller proportions as the general-domain component of the mixture. After tokenization, filtering, and shuffling, the final corpus contained approximately 508K sequences of length 1,024 tokens (Table~\ref{tab:data_composition}). Mixture selection was performed in a separate 130M pilot ablation before retraining the final 80\%/10\%/10\% configuration used for the main BioMamba checkpoints.

\begin{table}[t]
\centering
\caption{Continued pretraining data composition. C4 denotes the Colossal Clean Crawled Corpus.}
\label{tab:data_composition}
\begin{tabular}{@{}L{0.18\linewidth}L{0.42\linewidth}C{0.12\linewidth}C{0.12\linewidth}@{}}
\toprule
\textbf{Component} & \textbf{Source} & \textbf{Proportion} & \textbf{Sequences} \\
\midrule
PubMed & Abstracts indexed in MEDLINE (1,024-token sequences) & $\sim$80\% & $\sim$424K \\
C4 & General-domain web text & $\sim$10\% & $\sim$42K \\
Wikipedia & English Wikipedia & $\sim$10\% & $\sim$42K \\
\midrule
\textbf{Total} & & \textbf{100\%} & \textbf{$\sim$508K} \\
\bottomrule
\end{tabular}
\end{table}

\subsection{Training procedure}

Continued pretraining was performed for 3 epochs on the mixed PubMed/C4/Wikipedia corpus using AdamW \cite{Loshchilov2017DecoupledWD} optimization with weight decay 0.1, BF16 mixed precision, a maximum sequence length of 1,024 tokens, gradient clipping of 1.0, and random seed 42. To keep optimization roughly comparable across scales, we adjusted per-device microbatch size and gradient accumulation so that the effective batch size stayed approximately constant, at 240--256 sequences for the 130M--1.3B models and 192 sequences for the 2.7B model. This yielded 5,967--7,953 optimizer steps over the full 3-epoch training budget. Exponential moving average of model weights was enabled throughout training, and the checkpoint with the lowest validation loss was kept for downstream evaluation.

To mitigate catastrophic forgetting, we employed a conservative warmup--stable--decay (WSD) schedule together with layer-wise learning-rate decay, which updates lower layers more conservatively than higher layers using a layer-wise decay factor of 0.90 at 130M and 0.95 for the larger models. Peak learning rates decreased with model size, from $1.5\times10^{-6}$ at 130M to $2.0\times10^{-7}$ at 2.7B ($4.5\times10^{-7}$ for 370M, $3.0\times10^{-7}$ for 780M, and $2.5\times10^{-7}$ for 1.3B).

Shared and model-specific implementation details, including exact scheduler configuration, per-scale microbatch settings, hardware allocation, and memory usage, are reported in Supplementary Section~\ref{app:cpt_impl}. As shown later in the Results, training remained stable across all scales, with smooth decreases in validation loss and the best validation point occurring at or near the final step in every run.

\subsection{Evaluation setup}

\paragraph{Internal language-model evaluation.}
The primary language-model evaluation used fixed validation sets of 1,000 held-out sequences for PubMed \cite{canese2013pubmed} and C4 \cite{Dodge2021DocumentingLW}, drawn from the same data pools as continued pretraining but excluded from the training split, and tokenized with the shared tokenizer. For the Wikipedia held-out set, we used the Wikitext-103 raw split, referred to throughout this paper as the \emph{Wikipedia-style held-out} set for brevity, and tokenized it with the same shared tokenizer. This is the most controlled comparison in the study because the tokenizer is held constant across all internal evaluations and each benchmark split is fixed in advance.

\paragraph{Contextual comparison with public checkpoints.}
To place BioMamba alongside released biomedical and medically adapted checkpoints, we also evaluated public models on the same raw text using each model's native tokenizer. Because perplexity depends on tokenization, these results are contextual rather than strictly comparable and are interpreted only as secondary model-level evidence. Exact checkpoint sources are listed in Supplementary Section~\ref{app:eval_details}.

\subsection{Downstream tasks}

We evaluated downstream transfer to clinical text on two complementary supervised generation tasks constructed from de-identified discharge summaries in MIMIC-IV-Note v2.2 \cite{PhysioNet-mimic-iv-note-2.2} within the broader MIMIC-IV resource \cite{johnson2023mimiciv}. Note completion evaluates continuation ability: the model receives the first half of a discharge note and generates its continuation. Discharge summary generation evaluates summarization ability: the model receives structured admission sections and generates discharge sections. To avoid information leakage, splits were defined at the patient level, and quantitative evaluation used 500 randomly sampled held-out test notes per task. These clinical generation experiments were conducted at all five model scales. The two clinical generation settings are closely related to prior work on hospital-course summarization and discharge-note generation \cite{Adams2021WhatsIA,Hartman2022ADA,Searle2022DischargeSH,Aali2024ADA,Xu2024OverviewOT}. Full data construction, fine-tuning settings, and qualitative examples are provided in Supplementary Sections~\ref{app:mimic_details} and \ref{app:mimic_case_study}.

We also evaluated downstream transfer on biomedical yes/no question answering \cite{Jin2021BiomedicalQA} because it provides a simple and interpretable test of whether the model can use biomedical knowledge after continued pretraining. Supervised fine-tuning used a mixture of Biomedical Semantic Indexing and Question Answering (BioASQ) yes/no training examples from the BioASQ challenge series \cite{Tsatsaronis2015AnOO,Krithara2022BioASQQAAM,Nentidis2023BioASQ} aggregated from BioASQ 7B \cite{nentidis2020bioasq} and 13B \cite{Nentidis2025OverviewOB} (1,114 samples after class balancing) and a binary yes/no split derived from the PubMed Question Answering (PubMedQA) dataset \cite{jin2019pubmedqa} (800 training samples). Held-out evaluation was conducted on BioASQ 13B and the corresponding 200-sample binary PubMedQA test split. For both downstream settings, supervised fine-tuning updated all model parameters rather than using parameter-efficient adapters.

\subsection{Metrics and analysis}

For language modeling, we report cross-entropy-derived perplexity (PPL),
\begin{equation}
\mathrm{PPL} = \exp(\bar{\mathcal{L}}),
\end{equation}
where $\bar{\mathcal{L}}$ is the mean next-token cross-entropy on the evaluation set. Lower PPL indicates better language modeling. For downstream question answering, we report accuracy and macro-F1 score, where the macro-F1 score is the arithmetic mean of the class-specific yes and no F1 scores. Class-specific yes and no F1 scores are reported when helpful to show which class benefited most from continued pretraining.

For clinical text generation, note completion is treated as a completion task and discharge summary generation as a summarization task. For both, we report ROUGE-1, ROUGE-2, and ROUGE-L \cite{lin2004rouge}. ROUGE-$n$ measures the $n$-gram recall between the generated text and the reference:
\begin{equation}
\mathrm{ROUGE}\text{-}n = \frac{\sum_{s \in \mathcal{R}} \sum_{\mathrm{gram}_n \in s} \mathrm{Count}_{\mathrm{match}}(\mathrm{gram}_n)}{\sum_{s \in \mathcal{R}} \sum_{\mathrm{gram}_n \in s} \mathrm{Count}(\mathrm{gram}_n)},
\end{equation}
where $\mathcal{R}$ is the set of reference sentences and $\mathrm{Count}_{\mathrm{match}}$ counts the number of $n$-grams that appear in both the generated and reference texts. ROUGE-L uses the longest common subsequence (LCS) instead of fixed-length $n$-grams:
\begin{equation}
\mathrm{ROUGE}\text{-}L = \frac{(1 + \beta^2) \cdot P_{\mathrm{lcs}} \cdot R_{\mathrm{lcs}}}{R_{\mathrm{lcs}} + \beta^2 \cdot P_{\mathrm{lcs}}},
\end{equation}
where $P_{\mathrm{lcs}} = \mathrm{LCS}(X, Y) / |X|$ and $R_{\mathrm{lcs}} = \mathrm{LCS}(X, Y) / |Y|$ are LCS-based precision and recall, and $\beta$ is set to favor recall. Higher ROUGE scores indicate greater overlap with the reference text.

For downstream accuracy, we report 95\% Wilson binomial confidence intervals in the text when helpful to reflect the small size of the BioASQ test set ($n=82$).

\section{Results}
\label{sec:results}

\subsection{Biomedical language modeling}
\label{sec:biomed_lm}

\paragraph{Quantitative results.}
Under the tokenizer-controlled internal comparison, continued pretraining lowered PubMed perplexity at every model scale (Table~\ref{tab:base_vs_cpt}). 
%
Continued pretraining yielded consistent reductions in PubMed perplexity across all scales, dropping from 9.41 to 8.42 in the 130M model and scaling down to 5.28 in the 2.7B model.
The relative PubMed perplexity reduction ranged from 7\% to 11\% across scales, demonstrating that biomedical adaptation yields consistent gains even at the largest model size.
Wikipedia-style held-out perplexity also decreased at every scale, consistent with the inclusion of Wikipedia in the continued pretraining mixture. By contrast, C4 perplexity changed by at most about 1\%, so PubMed gains did not come at the cost of out-of-domain C4 performance before any task-specific fine-tuning.

\begin{table}[!htbp]
\centering
\caption{Internal perplexity comparison between base Mamba2 models and BioMamba. The base models are the public checkpoints used for initialization.}
\label{tab:base_vs_cpt}
\begin{tabular}{@{}L{0.22\linewidth}C{0.22\linewidth}C{0.24\linewidth}C{0.22\linewidth}@{}}
\toprule
\textbf{Model} & \textbf{PubMed PPL} $(\downarrow)$ & \textbf{Wikipedia PPL} $(\downarrow)$ & \textbf{C4 PPL} $(\downarrow)$ \\
\midrule
Mamba2-130M (base) & 9.41 & 23.09 & 26.58 \\
BioMamba-130M & \textbf{8.42} & \textbf{18.76} & 26.58 \\
\midrule
Mamba2-370M (base) & 7.41 & 16.17 & 20.06 \\
BioMamba-370M & \textbf{6.75} & \textbf{13.36} & \textbf{20.01} \\
\midrule
Mamba2-780M (base) & 6.61 & 13.54 & 17.32 \\
BioMamba-780M & \textbf{6.15} & \textbf{11.58} & 17.32 \\
\midrule
Mamba2-1.3B (base) & 6.12 & 11.91 & 15.59 \\
BioMamba-1.3B & \textbf{5.66} & \textbf{10.04} & \textbf{15.43} \\
\midrule
Mamba2-2.7B (base) & 5.67 & 10.28 & 13.99 \\
BioMamba-2.7B & \textbf{5.28} & \textbf{8.85} & \textbf{13.87} \\
\bottomrule
\end{tabular}
\end{table}

\begin{figure}[!htbp]
\centering
\safeincludegraphics[width=0.75\linewidth]{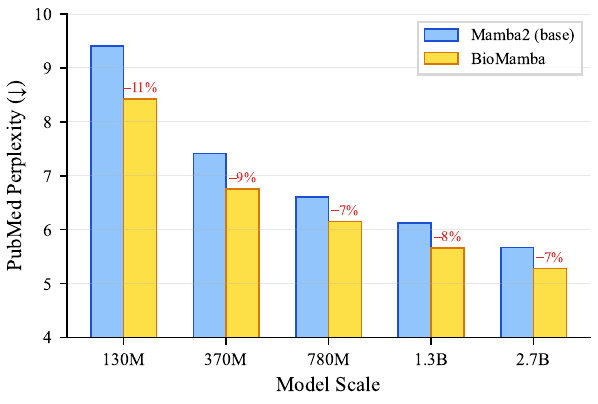}
\caption{PubMed perplexity comparison between base Mamba2 models and BioMamba across five model scales. Percentage labels indicate the relative perplexity reduction achieved by continued pretraining at each fixed model scale. The within-scale gap between each base model and its BioMamba counterpart is largest at 130M, indicating that the marginal benefit of continued pretraining is most pronounced for the smallest model.}
\label{fig:ppl_comparison}
\end{figure}

\paragraph{Case study.}
The smallest model, BioMamba-130M, shows the main pattern clearly. Continued pretraining lowered PubMed perplexity from 9.41 to 8.42 and Wikipedia-style held-out perplexity from 23.09 to 18.76, while C4 perplexity remained 26.58. This example shows that the mixed-corpus strategy improved biomedical language modeling without harming the general-language-modeling fluency measured on C4.

\paragraph{Contextual comparison with public checkpoints.}
Table~\ref{tab:external_comparison} reports raw-text perplexity for BioMamba alongside released biomedical and medically adapted checkpoints, each evaluated with its own native tokenizer. Because tokenizers, pretraining data, and adaptation protocols differ, we use these results only to situate BioMamba within the released-model landscape rather than as a controlled architecture-level comparison; within that contextual view, BioMamba-2.7B falls among the strongest raw-text perplexities in this set, while BioMamba-130M remains in a competitive range at much smaller scale.

\begin{table}[!htbp]
\centering
\caption{Raw-text perplexity comparison with released biomedical and medically adapted language models, each evaluated on the same text with its own native tokenizer.}
\label{tab:external_comparison}
\small
\setlength{\tabcolsep}{2pt}
\begin{tabular}{@{}L{0.27\linewidth}C{0.13\linewidth}C{0.19\linewidth}C{0.21\linewidth}C{0.15\linewidth}@{}}
\toprule
\textbf{Model} & \textbf{\# Params} & \textbf{PubMed PPL} $(\downarrow)$ & \textbf{Wikipedia PPL} $(\downarrow)$ & \textbf{C4 PPL} $(\downarrow)$ \\
\midrule
BioMamba-130M (ours) & 130M & 11.08 & 20.49 & 29.01 \\
BioGPT & 347M & 15.09 & 61.31 & 104.62 \\
Bio-Medical-Llama-3.2-1B & 1B & 14.46 & 22.54 & 36.42 \\
Gemma3-1B (fine-tuned) & $\sim$1B & 19.90 & 35.01 & 41.67 \\
BioGPT-Large & 1.5B & 10.00 & 35.83 & 50.79 \\
BioGPT-Large-PubMedQA & 1.5B & 9.26 & 43.54 & 71.96 \\
Meditron3-Gemma2-2B & 2B & 6.67 & 11.61 & 15.62 \\
BioMedLM & 2.7B & 25.27 & 100.24 & 95.74 \\
\textbf{BioMamba-2.7B (ours)} & \textbf{2.7B} & \textbf{6.52} & \textbf{9.71} & \textbf{14.91} \\
\bottomrule
\end{tabular}
\end{table}

\subsection{Out-of-domain evaluation and evaluation-budget stability}
\label{sec:ood_longctx}

To evaluate whether the 80\%/10\%/10\% recipe preserves general-language-modeling fluency beyond the in-domain perplexity benchmarks, we added two complementary analyses. First, we evaluated all five BioMamba checkpoints and their Mamba2 base counterparts on six out-of-domain multiple-choice benchmarks (LAMBADA, HellaSwag, ARC-Easy, ARC-Challenge, PIQA, OpenBookQA), scored by continuation log-likelihood. Across tasks and scales, BioMamba stays within $\pm 3$ percentage points of Mamba2 on every task with no systematic regression (Table~\ref{tab:ood_mcq_summary}); the per-scale mean change ranges only from $-0.51$ to $+0.58$ percentage points, which is consistent with the absence of catastrophic forgetting on broad commonsense reasoning, reading comprehension, and elementary-science benchmarks.

\begin{table}[!htbp]
\centering
\caption{Out-of-domain multiple-choice summary across six benchmarks. For each scale we report the mean BioMamba$-$Mamba2 accuracy difference averaged over the six tasks and the range. All per-task $\Delta$ lie inside $\pm 3$\,pp at every scale. The full accuracy grid is reported in Supplementary Table~\ref{tab:ood_mcq_full}.}
\label{tab:ood_mcq_summary}
\begin{tabular}{@{}C{0.15\linewidth}C{0.30\linewidth}C{0.45\linewidth}@{}}
\toprule
\textbf{Scale} & \textbf{Mean $\Delta$ (pp)} & \textbf{Per-task $\Delta$ range (pp)} \\
\midrule
130M & $-0.22$ & $[-2.10,\,+0.80]$ \\
370M & $+0.22$ & $[-1.60,\,+1.67]$ \\
780M & $-0.51$ & $[-1.20,\,+0.53]$ \\
1.3B & $+0.58$ & $[-0.33,\,+2.36]$ \\
2.7B & $+0.10$ & $[-0.67,\,+0.60]$ \\
\bottomrule
\end{tabular}
\end{table}

Second, we checked that the Section~\ref{sec:biomed_lm} perplexity numbers remain stable when the evaluation budget is increased from 1024 to 2048 and 4096 tokens. We therefore re-evaluated C4, Wikitext-103, and PubMed-medline at evaluation budgets of 1024, 2048, and 4096 tokens. Across all scales and budgets, values agree within $\pm 0.01$ PPL; Table~\ref{tab:longctx_ppl} therefore reports a representative single column rather than a three-column grid (the 2048 column is representative of 1024 and 4096 under this rounding). The Mamba2$\to$BioMamba deltas are $|\Delta\mathrm{PPL}| \le 0.17$ on C4, $-1.46$ to $-4.72$ PPL on Wikitext-103, and $-0.11$ to $-0.44$ PPL on PubMed-medline, with the Wikitext-103 gain largest at the smaller scales and decreasing monotonically with model size---consistent with the Section~\ref{sec:biomed_lm} finding that continued pretraining is especially helpful in the lower-cost regime.

\begin{table}[!htbp]
\centering
\caption{Extended-budget perplexity for Mamba2 base and BioMamba on C4, Wikitext-103, and PubMed-medline. Each numeric cell shows Mamba2$\to$BioMamba with the delta in parentheses.}
\label{tab:longctx_ppl}
\small
\setlength{\tabcolsep}{3pt}
\begin{tabular}{@{}C{0.08\linewidth}C{0.26\linewidth}C{0.30\linewidth}C{0.26\linewidth}@{}}
\toprule
\textbf{Scale} & \textbf{C4 PPL} & \textbf{Wikitext-103 PPL} & \textbf{PubMed PPL} \\
\midrule
130M & $22.42 \to 22.40$ ($-0.02$) & $23.42 \to \mathbf{18.70}$ ($-4.72$) & $11.67 \to 11.23$ ($-0.44$) \\
370M & $17.09 \to 17.05$ ($-0.04$) & $16.41 \to \mathbf{13.61}$ ($-2.80$) & $9.06 \to 8.80$ ($-0.26$) \\
780M & $14.84 \to 14.81$ ($-0.03$) & $13.66 \to \mathbf{11.62}$ ($-2.04$) & $7.99 \to 7.85$ ($-0.14$) \\
1.3B & $13.37 \to 13.25$ ($-0.12$) & $12.03 \to \mathbf{10.06}$ ($-1.97$) & $7.40 \to 7.26$ ($-0.14$) \\
2.7B & $12.10 \to 11.97$ ($-0.13$) & $10.36 \to \mathbf{8.90}$ ($-1.46$) & $6.80 \to 6.69$ ($-0.11$) \\
\bottomrule
\end{tabular}
\end{table}

\subsection{Clinical note completion and discharge summary generation on \mbox{MIMIC-IV}}

We next asked whether biomedical adaptation learned from PubMed would transfer to clinical documentation. We therefore evaluated the models on MIMIC-IV-Note v2.2, using the patient-level split and 500-sample held-out evaluation described in the Methods and in Supplementary Section~\ref{app:mimic_details}.

\paragraph{Clinical supervised fine-tuning.}
Starting from both the Mamba2 base checkpoints and the BioMamba checkpoints, we performed clinical supervised fine-tuning (SFT) on two generation tasks constructed from the discharge notes. Training data comprised up to 20{,}000 note-completion samples and approximately 12{,}893 discharge-generation samples (32{,}893 total; 90\% train, 10\% validation). Instruction masking was applied so that only response tokens contributed to the loss. BioMamba+SFT uses a lower learning rate and more training epochs than Mamba2+SFT because aggressive fine-tuning tends to overwrite the information acquired during continued pretraining; to verify that this design choice does not create an unfair comparison, we additionally ran a matched-budget control in which Mamba2 was fine-tuned with the exact BioMamba SFT recipe, and the BioMamba schedule did not close the gap (Supplementary Section~\ref{app:matched_budget}). We also found that validation loss was not a reliable proxy for downstream ROUGE in the BioMamba setting, which further motivates the conservative BioMamba SFT schedule. Full SFT data construction, hyperparameter configurations, and tuning details are provided in Supplementary Section~\ref{app:mimic_details}.

\paragraph{Evaluation tasks.}
We evaluated two clinical generation tasks that reflect different capabilities. In note completion, the model receives the first half of a discharge note and generates a 128-token continuation (greedy decoding), scored against the ground-truth second half by ROUGE-1, ROUGE-2, and ROUGE-L. In discharge summary generation, the model receives structured admission sections and generates discharge sections, again scored by ROUGE.

\paragraph{Results.}
Table~\ref{tab:mimic_main} presents MIMIC-IV ROUGE across all five scales, with the \texttt{Mamba2+SFT} / \texttt{BioMamba+SFT} label convention of Table~\ref{tab:sft_biomamba_only}. For the 2.7B row we additionally report, in Supplementary Section~\ref{app:bootstrap_ci}, (i)~per-arm 95\% percentile-bootstrap confidence intervals for ROUGE-1 and ROUGE-L (Supplementary Table~\ref{tab:bootstrap_abs}) and (ii)~paired-bootstrap significance on per-sample ROUGE-1 \emph{differences} (Supplementary Table~\ref{tab:paired_bootstrap}), both with 1000 resamples over the 500 test samples. Three patterns emerge.

First, clinical SFT is the dominant driver of generation quality: at the 130M, 370M, 780M, and 1.3B scales where base Mamba2 rows are reported, both Mamba2+SFT and BioMamba+SFT substantially outperform the corresponding base models on every ROUGE metric.

Second, BioMamba+SFT consistently matches or outperforms Mamba2+SFT on ROUGE-1 and ROUGE-L at every scale on both completion and discharge generation, with the single exception of a 0.01-point ROUGE-2 gap on 1.3B discharge (3.70 vs.\ 3.69). The advantage is most pronounced on discharge generation at 130M (ROUGE-1 9.74 vs.\ 8.79, $\Delta$=+0.95). At 2.7B, paired-bootstrap tests on per-sample ROUGE-1 differences give $P(\Delta > 0) = 1.00$ for completion and $P(\Delta > 0) = 0.997$ for discharge, so the advantage is not within test-sample noise.

Third, within each arm MIMIC-IV ROUGE improves with scale up to 1.3B (BioMamba+SFT reaches 8.11 / 10.11 on completion / discharge at 1.3B) and is slightly lower at 2.7B (7.57 / 9.39); the BioMamba+SFT advantage over Mamba2+SFT nonetheless persists at every scale, so the effect of continued pretraining is not absorbed by scaling on this dataset.

\begin{table}[!htbp]
\centering
\caption{MIMIC-IV clinical generation results (ROUGE \%, 500 test samples per task). Boldface indicates the best result within each model size. Comp = note completion; Disch = discharge summary generation. R-$1/2/L$ = Rouge-$1/2/L$. The 2.7B block omits a base-only row because Mamba2+SFT is the appropriate fairness baseline at that scale. Per-arm 95\% percentile-bootstrap CIs for R-1 and R-L at 2.7B (Supplementary Table~\ref{tab:bootstrap_abs}) and paired-bootstrap significance on per-sample ROUGE-1 differences at 2.7B (Supplementary Table~\ref{tab:paired_bootstrap}) are both in Supplementary Section~\ref{app:bootstrap_ci}.}
\label{tab:mimic_main}
\begin{tabular}{@{}L{0.22\linewidth}*{6}{S[table-format=2.2]}@{}}
\toprule
& \multicolumn{3}{c}{\textbf{Note completion}} & \multicolumn{3}{c}{\textbf{Discharge generation}} \\
\cmidrule(lr){2-4}\cmidrule(l){5-7}
\textbf{Model} & {\textbf{R-1} $\uparrow$} & {\textbf{R-2} $\uparrow$} & {\textbf{R-L} $\uparrow$} & {\textbf{R-1} $\uparrow$} & {\textbf{R-2} $\uparrow$} & {\textbf{R-L} $\uparrow$} \\
\midrule
\multicolumn{7}{@{}l}{\textit{130M}} \\
Mamba2 & 4.63 & 0.80 & 3.43 & 4.83 & 0.35 & 3.56 \\
Mamba2+SFT & 6.84 & 2.36 & 4.95 & 8.79 & 2.76 & 6.30 \\
BioMamba+SFT & \bfseries 7.07 & \bfseries 2.45 & \bfseries 5.10 & \bfseries 9.74 & \bfseries 3.33 & \bfseries 6.96 \\
\midrule
\multicolumn{7}{@{}l}{\textit{370M}} \\
Mamba2 & 5.04 & 0.97 & 3.69 & 4.50 & 0.33 & 3.31 \\
Mamba2+SFT & 7.85 & 3.04 & 5.62 & 8.96 & 2.69 & 6.19 \\
BioMamba+SFT & \bfseries 7.94 & \bfseries 3.20 & \bfseries 5.69 & \bfseries 9.22 & \bfseries 3.00 & \bfseries 6.39 \\
\midrule
\multicolumn{7}{@{}l}{\textit{780M}} \\
Mamba2 & 5.46 & 1.17 & 3.86 & 5.22 & 0.37 & 3.59 \\
Mamba2+SFT & \bfseries 7.90 & 3.24 & 5.73 & 8.86 & 2.73 & 6.14 \\
BioMamba+SFT & \bfseries 7.90 & \bfseries 3.32 & \bfseries 5.78 & \bfseries 9.10 & \bfseries 2.97 & \bfseries 6.30 \\
\midrule
\multicolumn{7}{@{}l}{\textit{1.3B}} \\
Mamba2 & 5.61 & 1.28 & 3.95 & 5.33 & 0.42 & 3.68 \\
Mamba2+SFT & 7.93 & 3.28 & 5.76 & 9.99 & \bfseries 3.70 & \bfseries 7.04 \\
BioMamba+SFT & \bfseries 8.11 & \bfseries 3.33 & \bfseries 5.89 & \bfseries 10.11 & 3.69 & \bfseries 7.04 \\
\midrule
\multicolumn{7}{@{}l}{\textit{2.7B}} \\
Mamba2 & 5.55 & 1.19 & 3.85 & 7.04	 & 0.80	 & 4.71 \\
Mamba2+SFT & 7.16 & 2.45 & 4.98 & 8.87 & 2.76 & 6.21 \\
BioMamba+SFT & \bfseries 7.57 & \bfseries 2.83 & \bfseries 5.30 & \bfseries 9.39 & \bfseries 3.01 & \bfseries 6.46 \\
\bottomrule
\end{tabular}
\end{table}

\paragraph{Representative case studies.}
The ROUGE improvements corresponded to clear qualitative differences in clinical generation. Table~\ref{tab:case_completion_main} shows a representative note-completion example from the 1.3B model family. The base Mamba2 model incorrectly propagated the cardiac abbreviation ``RRR'' to unrelated organ systems, producing clinically implausible output. Mamba2+SFT corrected the basic clinical register, whereas BioMamba+SFT produced the most complete continuation, including pertinent negatives and a natural transition to the discharge physical-examination section.

Table~\ref{tab:case_summarization_main} shows a representative discharge-summary generation example after vascular surgery. Here the base Mamba2 model largely repeated procedural details, Mamba2+SFT generated only a partial discharge framework, and BioMamba+SFT moved directly to a structured discharge-medication section with a clinically coherent regimen. These examples indicate that the quantitative gains reflect better completion and summarization behavior rather than only small changes in lexical overlap. Additional qualitative cases, including multi-panel laboratory continuation, medication-reconciliation continuation, Hospital Course generation, and concise discharge instructions, are provided in Supplementary Section~\ref{app:mimic_case_study}.

\begin{table}[!htbp]
\centering
\caption{Representative note-completion case study from the 1.3B model family. The input ends with ``\ldots HEART: RRR, no MRG, nl''. Each non-reference row contains a short \emph{excerpt} from the model's generation (before the em-dash) and a one-line \emph{author reading} of that excerpt. Full raw outputs for this and three additional examples are in Supplementary Section~\ref{app:mimic_case_study}.}
\label{tab:case_completion_main}
\small
\setlength{\tabcolsep}{4pt}
\begin{tabularx}{\linewidth}{@{}L{0.19\linewidth}Y C{0.08\linewidth}@{}}
\toprule
\textbf{Source} & \textbf{Excerpt (before em-dash) + author reading (after em-dash)} & \textbf{R-1} \\
\midrule
Reference & S1-S2. LUNGS: CTA bilat, no ronchi/rales/wheezes\ldots\ ABDOMEN: Midline surgical scar. Right sided PD catheter\ldots\ EXTREMITIES: WWP, no edema, 2+ peripheral pulses. NEURO: Awake, A\&Ox3, CNs II-XII grossly intact. \emph{(reference text; no reading.)} & --- \\
\midrule
Mamba2 & ``LUNGS: RRR, no rales; ABDOMEN: RRR, no tenderness; URINARY: RRR\ldots\ DIAPER: RRR'' --- \emph{incorrectly applies the cardiac rhythm descriptor ``RRR'' to unrelated organ systems.} & 2.73 \\
\midrule
Mamba2+SFT & ``S1-S2; LUNGS: CTAB, no r/rh/wh\ldots\ ABDOMEN: Soft, NTND\ldots\ EXTREMITIES: WWP'' --- \emph{clinically plausible continuation with standard abbreviations.} & 3.65 \\
\midrule
\textbf{BioMamba+SFT} & ``S1-S2; LUNGS: CTAB, no wheezes, rales, rhonchi\ldots\ ABDOMEN: Soft, NTND, \textbf{no rebound or guarding}\ldots\ NEURO: A\&Ox3, moving all four extremities\ldots\ \textbf{Physical Exam on Discharge:}'' --- \emph{adds pertinent negatives and transitions to the discharge physical-examination section.} & \textbf{3.84} \\
\bottomrule
\end{tabularx}
\end{table}

\begin{table}[!htbp]
\centering
\caption{Representative discharge-summary generation case study from the 1.3B model family. Input: Service: SURGERY; Chief Complaint: Foot Ulcer; Procedure: Ultrasound-guided access to the left common femoral artery, selective catheterization of the right superficial femoral artery\ldots\ Cell text is a \emph{short excerpt} from the model's generation (verbatim tokens in quotes) followed by an \emph{author reading} (in italics). Full raw outputs for this and three additional examples are in Supplementary Section~\ref{app:mimic_case_study}.}
\label{tab:case_summarization_main}
\small
\setlength{\tabcolsep}{4pt}
\begin{tabularx}{\linewidth}{@{}L{0.19\linewidth}Y C{0.08\linewidth}C{0.08\linewidth}@{}}
\toprule
\textbf{Source} & \textbf{Excerpt + author reading (italicized)} & \textbf{R-1} & \textbf{R-2} \\
\midrule
Reference & DISCHARGE MEDICATIONS: 1.\ Clopidogrel 75\,mg PO DAILY 2.\ Sulfameth/Trimethoprim DS 2 TAB PO BID\ldots\ \emph{(reference text; no reading.)} & --- & --- \\
\midrule
Mamba2 & ``8 x 80\,mm Innova stent post dilated with a 7\,mm balloon'' (procedure phrasing repeated) --- \emph{repeats procedure details and fails to transition to discharge content.} & 10.85 & 0.68 \\
\midrule
Mamba2+SFT & \emph{Generates a partial discharge framework and follow-up surgical description, but no explicit medication list. (Excerpt omitted here for space; see Supplementary Section~\ref{app:mimic_case_study}.)} & 17.87 & 2.08 \\
\midrule
\textbf{BioMamba+SFT} & ``\textbf{DISCHARGE MEDICATIONS:} Aspirin 81\,mg, Atorvastatin 40\,mg, Metoprolol Tartrate 50\,mg\ldots'' --- \emph{concisely summarizes the procedure and moves directly to a structured medication regimen consistent with post-vascular surgery care.} & \textbf{23.05} & \textbf{3.75} \\
\bottomrule
\end{tabularx}
\end{table}

These results show that biomedical knowledge acquired during PubMed continued pretraining transfers to clinical text generation. The consistent advantage of BioMamba+SFT over Mamba2+SFT suggests that literature-based continued pretraining adds value even when clinical SFT data are available, although learning-rate tuning remains important during fine-tuning.

\paragraph{Output diagnostics.}
To complement the ROUGE comparison with quantitative summaries of generation behavior, we computed standard output diagnostics on 1.3B discharge generation under nucleus-0.9 and nucleus-0.95 decoding (Table~\ref{tab:mimic_diagnostics}): average length, distinct-1 and distinct-2 token-type ratios, and a repeated-4-gram rate. Greedy decoding remains the apples-to-apples primary setting for Table~\ref{tab:mimic_main}; we report diversity and repetition diagnostics only for the higher-entropy nucleus settings, where these metrics are informative. Under nucleus-0.95, a standard deployment setting for open-ended clinical text, BioMamba has slightly better diversity and a lower repeated-4-gram rate than Mamba2; under nucleus-0.9 the two arms are close. A lightweight section-header F1 structure metric is reported in Supplementary Table~\ref{tab:section_header_f1}.

\begin{table}[!htbp]
\centering
\caption{MIMIC-IV discharge output diagnostics at 1.3B under nucleus-0.9 and nucleus-0.95 decoding: average generated length, distinct-1 and distinct-2 token-type ratios, and the repeated-4-gram rate. Greedy decoding used in Table~\ref{tab:mimic_main} is omitted from this table.}
\label{tab:mimic_diagnostics}
\small
\setlength{\tabcolsep}{4pt}
\begin{tabular}{@{}L{0.18\linewidth}L{0.18\linewidth}C{0.12\linewidth}C{0.11\linewidth}C{0.11\linewidth}C{0.15\linewidth}@{}}
\toprule
\textbf{Model} & \textbf{Decoding} & \textbf{Avg length} & \textbf{Dist-1} & \textbf{Dist-2} & \textbf{Rep-4-gram} \\
\midrule
Mamba2+SFT & nucleus-0.9 & 74.6 & 0.730 & 0.876 & 0.071 \\
BioMamba+SFT & nucleus-0.9 & 74.8 & 0.719 & 0.865 & 0.077 \\
\midrule
Mamba2+SFT & nucleus-0.95 & 74.1 & 0.770 & 0.914 & 0.040 \\
\textbf{BioMamba+SFT} & nucleus-0.95 & 74.6 & \textbf{0.777} & \textbf{0.918} & \textbf{0.036} \\
\bottomrule
\end{tabular}
\end{table}

\paragraph{Factuality proxy analysis.}
To move beyond ROUGE as a surface-lexical metric and provide a grounding-oriented check of clinical generation, we added an NLI-ensemble factuality proxy on 2.7B MIMIC discharge generations (Table~\ref{tab:nli_factuality}). Using two MNLI checkpoints (\texttt{roberta-large-mnli} and \texttt{microsoft/deberta-v2-xlarge-mnli}), a generation is flagged as contradictory only when both checkpoints independently output a contradiction probability above $0.7$. We scored entailment of each generated discharge section against the input admission text only. Across all six tested cells the contradiction rate lies in the $7.0$--$9.4\%$ range, with Wilson 95\% CIs overlapping substantially between arms at every decoding. Macro-averaged across the three decoding strategies, BioMamba ($7.4\%$) is below Mamba2 ($8.1\%$); under nucleus-0.95 sampling Mamba2 rises to $9.4\%$ while BioMamba stays at $7.0\%$---a $2.4$~pp gap in BioMamba's favour. Continued pretraining therefore does not introduce systematic source-contradictory output relative to the base model, and is more stable than the base model under higher-entropy sampling. We present this analysis as a factuality proxy rather than a clinician audit and treat safety, calibration, and workflow readiness as out-of-scope for this study (see Section~\ref{sec:discussion}); the NLI protocol and the schema of the auto-flagged clinician-review candidate list are in Supplementary Section~\ref{app:nli_protocol}.

\begin{table}[!htbp]
\centering
\caption{NLI-ensemble factuality proxy on 2.7B MIMIC discharge generation ($n=500$ per cell; two MNLI checkpoints, \texttt{roberta-large-mnli} and \texttt{microsoft/deberta-v2-xlarge-mnli}). M2+SFT = Mamba2+SFT; BM+SFT = BioMamba+SFT. Contradiction rate is computed against the input admission text only, lower is better. 95\% confidence intervals are Wilson score intervals.}
\label{tab:nli_factuality}
\small
\setlength{\tabcolsep}{5pt}
\begin{tabular}{@{}L{0.14\linewidth}L{0.18\linewidth}C{0.22\linewidth}C{0.30\linewidth}@{}}
\toprule
\textbf{2.7B arm} & \textbf{Decoding} & \textbf{Contradictions / $n$} & \textbf{Rate ($\downarrow$) [Wilson 95\% CI]} \\
\midrule
\multirow{4}{*}{M2+SFT} & greedy       & 35 / 500 & 7.0\% $[5.1,\,9.6]$ \\
                        & nucleus-0.9  & 39 / 500 & 7.8\% $[5.8,\,10.5]$ \\
                        & nucleus-0.95 & 47 / 500 & 9.4\% $[7.1,\,12.3]$ \\
\cmidrule(l){2-4}
                        & macro-avg    & ---      & 8.1\% \\
\midrule
\multirow{4}{*}{BM+SFT} & greedy       & 37 / 500 & 7.4\% $[5.4,\,10.0]$ \\
                        & nucleus-0.9  & 39 / 500 & 7.8\% $[5.8,\,10.5]$ \\
                        & nucleus-0.95 & 35 / 500 & 7.0\% $[5.1,\,9.6]$ \\
\cmidrule(l){2-4}
                        & macro-avg    & ---      & 7.4\% \\
\bottomrule
\end{tabular}
\end{table}

\paragraph{Failure modes.}
The remaining errors are consistent with the behaviors already visible in the qualitative examples. For MIMIC, the main residual issues are occasional repetition under greedy decoding, occasional truncation of the target section when the 128-token budget is reached before the discharge medication list completes, and incomplete coverage of medication-level detail in a subset of generations; these are directly reflected in the nucleus-0.95 diagnostic improvements and in the qualitative case studies in Supplementary Section~\ref{app:mimic_case_study}. For the question-answering tasks, most BioASQ errors at larger scales are shared between Mamba2+SFT and BioMamba+SFT: at 780M, every item BioMamba+SFT gets wrong is also missed by Mamba2+SFT, and at 1.3B the overlap is 11 of 12 (see Supplementary Section~\ref{app:wrong_index}). Residual disagreements at larger scales are therefore confined to a handful of genuinely hard items rather than indicating a regression attributable to continued pretraining.

\subsection{Contextual comparison with external models on \mbox{MIMIC-IV}}

Table~\ref{tab:mimic_external} compares BioMamba's clinical generation results with seven public biomedical or medically adapted language models on the same 500 MIMIC-IV test samples under identical greedy decoding. The external baselines were not given MIMIC-specific supervised fine-tuning, so this is a contextual rather than task-matched comparison; caveats are summarized in the caption.

\begin{table}[!htbp]
\centering
\caption{Comparison of BioMamba with external biomedical and medically adapted language models on MIMIC-IV clinical generation (ROUGE \%, 500 test samples). Comp = note completion; Disch = discharge summary generation; R-$1/2/L$ = Rouge-$1/2/L$.}
\label{tab:mimic_external}
\footnotesize
\setlength{\tabcolsep}{3pt}
\begin{tabular}{@{}L{0.25\linewidth}C{0.13\linewidth}C{0.07\linewidth}*{6}{C{0.07\linewidth}}@{}}
\toprule
\multirow{2}{*}{\textbf{Model}} & \multirow{2}{*}{\textbf{Architecture}} & \multirow{2}{*}{\textbf{Params}} & \multicolumn{3}{c}{\textbf{Note completion}} & \multicolumn{3}{c}{\textbf{Discharge generation}} \\
\cmidrule(lr){4-6}\cmidrule(l){7-9}
 & & & \textbf{R-1} $\uparrow$ & \textbf{R-2} $\uparrow$ & \textbf{R-L} $\uparrow$ & \textbf{R-1} $\uparrow$ & \textbf{R-2} $\uparrow$ & \textbf{R-L} $\uparrow$ \\
\midrule
BioGPT & GPT-2 & 347M & 1.20 & 0.16 & 1.01 & 1.53 & 0.09 & 1.26 \\
BioMedLM & GPT-2 & 2.7B & 1.58 & 0.16 & 1.17 & 2.18 & 0.13 & 1.59 \\
BioGPT-Large & GPT-2 & 1.5B & 3.85 & 0.54 & 2.81 & 3.20 & 0.20 & 2.42 \\
Gemma3-1B (fine-tuned) & Gemma3 & 1B & 4.18 & 0.65 & 2.83 & 5.55 & 0.42 & 3.77 \\
BioGPT-Large-PubMedQA & GPT-2 & 1.5B & 4.75 & 0.75 & 3.04 & 5.28 & 0.41 & 3.30 \\
Bio-Medical-Llama-3.2-1B & Llama 3.2 & 1B & 4.80 & 0.82 & 3.19 & 5.58 & 0.43 & 3.70 \\
Meditron3-Gemma2-2B & Gemma2 & 2B & 5.13 & 0.89 & 3.37 & 6.04 & 0.55 & 3.88 \\
\midrule
\textbf{BioMamba-130M+SFT} & \textbf{Mamba2} & \textbf{130M} & \textbf{7.07} & \textbf{2.45} & \textbf{5.10} & \textbf{9.74} & \textbf{3.33} & \textbf{6.96} \\
\textbf{BioMamba-1.3B+SFT} & \textbf{Mamba2} & \textbf{1.3B} & \textbf{8.11} & \textbf{3.33} & \textbf{5.89} & \textbf{10.11} & \textbf{3.69} & \textbf{7.04} \\
\bottomrule
\end{tabular}
\end{table}

Within this contextual setup, the task-specific BioMamba rows lie above the untuned external baselines on both MIMIC tasks. We interpret this only as evidence that BioMamba is competitive relative to released checkpoints in the current landscape, not as a controlled superiority claim, because the comparison does not isolate architecture, tokenizer, pretraining data, or task-adaptation protocol. The observed gap should therefore be read descriptively rather than causally.

\subsection{Biomedical question answering on BioASQ}

BioASQ yes/no question answering complements the MIMIC experiments because it tests literature-oriented biomedical reasoning rather than clinical document generation. Downstream performance after supervised fine-tuning, together with the corresponding Mamba2+SFT baseline accuracies and gains, is summarized in Table~\ref{tab:sft_biomamba_only}. BioASQ gains from continued pretraining are scale-dependent and non-monotonic: the 370M row is slightly below the Mamba2+SFT baseline, 130M, 780M, and 1.3B are positive, and 2.7B ties the baseline. Because the BioASQ yes/no test split contains only 82 items, a single-question difference corresponds to 1.22 percentage points, which matches the magnitude of the 370M and 2.7B deltas; a 3-seed reproduction across all five scales (seeds $\{13, 42, 97\}$; Supplementary Table~\ref{tab:multiseed_qa}) gives $\sigma \le 0.021$ on BioASQ and $\sigma \le 0.032$ on PubMedQA, and an item-level error analysis (Supplementary Table~\ref{tab:wrong_index_bioasq}) shows that BioMamba fixes more BioASQ items than it breaks at every scale while most residual errors are shared between the two arms. In contrast to the non-monotonic BioASQ pattern, PubMedQA gains from continued pretraining are positive at every scale. Within the BioMamba series itself, performance improved overall with model scale: BioASQ accuracy increased from 68.29\% at 130M to 90.24\% at 2.7B, and the macro-F1 score increased from 0.634 to 0.890. The strongest model, BioMamba-2.7B, achieved 90.24\% accuracy (74/82; 95\% Wilson confidence interval, 81.9\% to 95.0\%) and a macro-F1 score of 0.890. Together with the internal perplexity and MIMIC findings, these results indicate that gains from biomedical language modeling transfer across biomedical literature and clinical text rather than remaining confined to perplexity.

\begin{table}[!htbp]
\centering
\caption{Downstream question-answering results across model scales. M2+SFT = Mamba2+SFT; BM+SFT = BioMamba+SFT. Each M2+SFT and BM+SFT cell reports raw correct/total counts together with accuracy. Gain = BM+SFT $-$ M2+SFT in percentage points; F1 is the BM+SFT macro-F1. Test-set sizes: BioASQ $n=82$; PubMedQA $n=200$. The corresponding 95\% Wilson binomial confidence intervals for the BM+SFT cells are in Supplementary Table~\ref{tab:sft_qa_raw_ci}; three-seed reproduction means $\pm$ standard deviations are in Supplementary Table~\ref{tab:multiseed_qa}.}
\label{tab:sft_biomamba_only}
\small
\setlength{\tabcolsep}{4pt}
\resizebox{0.97\linewidth}{!}{%
\begin{tabular}{@{}C{0.08\linewidth}*{4}{C{0.10\linewidth}}*{4}{C{0.10\linewidth}}@{}}
\toprule
 & \multicolumn{4}{c}{\textbf{BioASQ}} & \multicolumn{4}{c}{\textbf{PubMedQA}} \\
\cmidrule(lr){2-5}\cmidrule(l){6-9}
\textbf{Scale} & \textbf{M2+SFT} & \textbf{BM+SFT} & \textbf{Gain} & \textbf{F1} & \textbf{M2+SFT} & \textbf{BM+SFT} & \textbf{Gain} & \textbf{F1} \\
\midrule
130M & \makecell{44/82\\(53.66\%)} & \makecell{56/82\\(68.29\%)} & $+14.63$ & 0.634 & \makecell{112/200\\(56.00\%)} & \makecell{126/200\\(63.00\%)} & $+7.00$ & 0.661 \\
370M & \makecell{65/82\\(79.27\%)} & \makecell{64/82\\(78.05\%)} & $-1.22$  & 0.721 & \makecell{131/200\\(65.50\%)} & \makecell{132/200\\(66.00\%)} & $+0.50$ & 0.690 \\
780M & \makecell{64/82\\(78.05\%)} & \makecell{65/82\\(79.27\%)} & $+1.22$  & 0.778 & \makecell{129/200\\(64.50\%)} & \makecell{135/200\\(67.50\%)} & $+3.00$ & 0.711 \\
1.3B & \makecell{68/82\\(82.93\%)} & \makecell{70/82\\(85.37\%)} & $+2.44$  & 0.831 & \makecell{130/200\\(65.00\%)} & \makecell{140/200\\(70.00\%)} & $+5.00$ & 0.740 \\
\textbf{2.7B} & \makecell{74/82\\(90.24\%)} & \makecell{\textbf{74/82}\\(\textbf{90.24\%})} & $+0.00$ & \textbf{0.890} & \makecell{141/200\\(70.50\%)} & \makecell{\textbf{146/200}\\(\textbf{73.00\%})} & $+2.50$ & \textbf{0.773} \\
\bottomrule
\end{tabular}
}
\end{table}

\begin{figure}[!htbp]
\centering
\safeincludegraphics[width=0.75\linewidth]{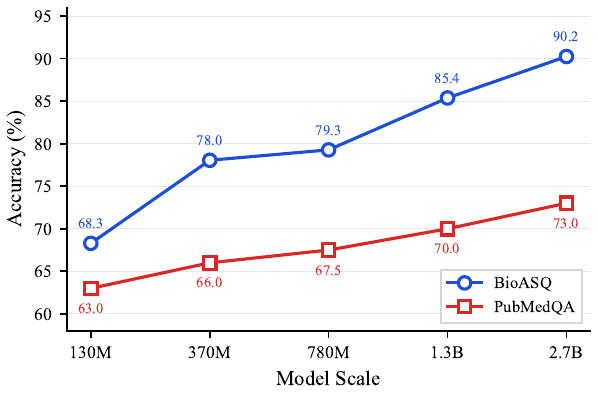}
\caption{Downstream question-answering accuracy as a function of model scale. Both BioASQ and PubMedQA accuracy improve consistently from 130M to 2.7B parameters.}
\label{fig:qa_scaling}
\end{figure}

\subsection{Biomedical question answering on PubMedQA}

Binary PubMedQA was the third evaluation task. Relative to the corresponding Mamba2+SFT baselines, BioMamba improved PubMedQA accuracy at every scale, and performance also improved overall with model scale within the BioMamba series (Table~\ref{tab:sft_biomamba_only}). PubMedQA accuracy increased from 63.00\% at 130M to 73.00\% at 2.7B, and the macro-F1 score increased from 0.661 to 0.773. The strongest model, BioMamba-2.7B, achieved 73.00\% accuracy (146/200; 95\% confidence interval, 66.5\% to 78.7\%) and a macro-F1 score of 0.773.

\subsection{Ablation studies}

\paragraph{Data-mixture ablation.}
The first ablation examined the composition of the continued pretraining corpus. As shown in Table~\ref{tab:mixing_full}, PubMed-only continued pretraining improved in-domain perplexity but worsened both Wikipedia and C4 perplexity, consistent with forgetting. Most mixed-corpus settings retained the PubMed gain, but many still weakened at least one held-out benchmark. The selected 10\% C4 + 10\% Wikipedia setting was the most balanced configuration: it achieved the lowest PubMed perplexity, the largest Wikipedia improvement, and essentially unchanged C4 perplexity. These results support the final 80\% PubMed, 10\% C4, and 10\% Wikipedia mixture used for the main BioMamba checkpoints. A complementary 370M PubMed-only control, reported in Supplementary Table~\ref{tab:mixing_370m}, confirms that this mixture remains load-bearing at a larger scale: PubMed-only continued pretraining worsens Wikipedia-style held-out perplexity by 69\% and C4 perplexity by 15\%, and is also slightly worse on held-out PubMed itself (9.32 $\to$ 9.91), consistent with the 130M pilot where the 80\%/10\%/10\% mixture already matched PubMed-only on PubMed (8.41 vs.\ 8.47); the gap is larger at 370M, indicating that domain-only continued pretraining incurs a larger forgetting penalty at scale even on in-domain text.

\begin{table}[!htbp]
\centering
\caption{Full continued pretraining data-mixing ablation for the 130M pilot selection run. Negative $\Delta$ values indicate lower perplexity than the base model.}
\label{tab:mixing_full}
\footnotesize
\setlength{\tabcolsep}{1.5pt}
\begin{tabular}{@{}L{0.26\linewidth}C{0.05\linewidth}C{0.11\linewidth}C{0.10\linewidth}C{0.09\linewidth}C{0.11\linewidth}C{0.11\linewidth}C{0.05\linewidth}C{0.05\linewidth}@{}}
\toprule
\makecell[t]{\textbf{Configuration}\\\strut} & \makecell[t]{\textbf{C4}\\\textbf{(\%)}} & \makecell[t]{\textbf{Wikipedia}\\\textbf{(\%)}} & \makecell[t]{\textbf{PubMed}\\\textbf{PPL}} & \makecell[t]{\textbf{PubMed}\\\textbf{$\Delta$}} & \makecell[t]{\textbf{Wikipedia}\\\textbf{PPL}} & \makecell[t]{\textbf{Wikipedia}\\\textbf{$\Delta$}} & \makecell[t]{\textbf{C4}\\\textbf{PPL}} & \makecell[t]{\textbf{C4}\\\textbf{$\Delta$}} \\
\midrule
No continued pretraining & 0 & 0 & 9.39 & --- & 23.29 & --- & 26.51 & --- \\
PubMed only & 0 & 0 & 8.47 & $\downarrow$9.8\% & 24.50 & $\uparrow$5.2\% & 27.90 & $\uparrow$5.2\% \\
\midrule
C4 only, 5\% & 5 & 0 & 8.48 & $\downarrow$9.7\% & 24.03 & $\uparrow$3.2\% & 26.53 & $\uparrow$0.1\% \\
C4 only, 10\% & 10 & 0 & 8.99 & $\downarrow$4.2\% & 25.28 & $\uparrow$8.5\% & 27.80 & $\uparrow$4.9\% \\
C4 only, 20\% & 20 & 0 & 8.48 & $\downarrow$9.7\% & 24.19 & $\uparrow$3.9\% & 26.34 & $\downarrow$0.7\% \\
C4 only, 30\% & 30 & 0 & 8.49 & $\downarrow$9.6\% & 24.25 & $\uparrow$4.1\% & 26.29 & $\downarrow$0.8\% \\
\midrule
Wikipedia only, 10\% & 0 & 10 & 8.48 & $\downarrow$9.7\% & 23.79 & $\uparrow$2.2\% & 27.07 & $\uparrow$2.1\% \\
Wikipedia only, 20\% & 0 & 20 & 8.49 & $\downarrow$9.5\% & 23.83 & $\uparrow$2.3\% & 27.07 & $\uparrow$2.1\% \\
Wikipedia only, 30\% & 0 & 30 & 8.50 & $\downarrow$9.5\% & 23.84 & $\uparrow$2.4\% & 27.08 & $\uparrow$2.2\% \\
\midrule
5\% C4 + 5\% Wikipedia & 5 & 5 & 8.48 & $\downarrow$9.7\% & 23.75 & $\uparrow$2.0\% & 26.50 & $\downarrow$0.0\% \\
5\% C4 + 10\% Wikipedia & 5 & 10 & 8.49 & $\downarrow$9.6\% & 23.76 & $\uparrow$2.0\% & 26.50 & $\downarrow$0.0\% \\
10\% C4 + 5\% Wikipedia & 10 & 5 & 8.49 & $\downarrow$9.6\% & 23.78 & $\uparrow$2.1\% & 26.42 & $\downarrow$0.3\% \\
\textbf{10\% C4 + 10\% Wikipedia} & \textbf{10} & \textbf{10} & \textbf{8.41} & $\downarrow$\textbf{10.4\%} & \textbf{18.95} & $\downarrow$\textbf{18.6\%} & \textbf{26.51} & \textbf{0.0\%} \\
10\% C4 + 20\% Wikipedia & 10 & 20 & 8.50 & $\downarrow$9.5\% & 23.76 & $\uparrow$2.0\% & 26.45 & $\downarrow$0.2\% \\
15\% C4 + 15\% Wikipedia & 15 & 15 & 8.50 & $\downarrow$9.5\% & 23.77 & $\uparrow$2.1\% & 26.39 & $\downarrow$0.4\% \\
20\% C4 + 10\% Wikipedia & 20 & 10 & 8.49 & $\downarrow$9.5\% & 23.75 & $\uparrow$2.0\% & 26.34 & $\downarrow$0.6\% \\
20\% C4 + 20\% Wikipedia & 20 & 20 & 8.50 & $\downarrow$9.5\% & 23.70 & $\uparrow$1.8\% & 26.36 & $\downarrow$0.6\% \\
\bottomrule
\end{tabular}
\end{table}

\begin{figure}[!htbp]
\centering
\safeincludegraphics[width=\linewidth]{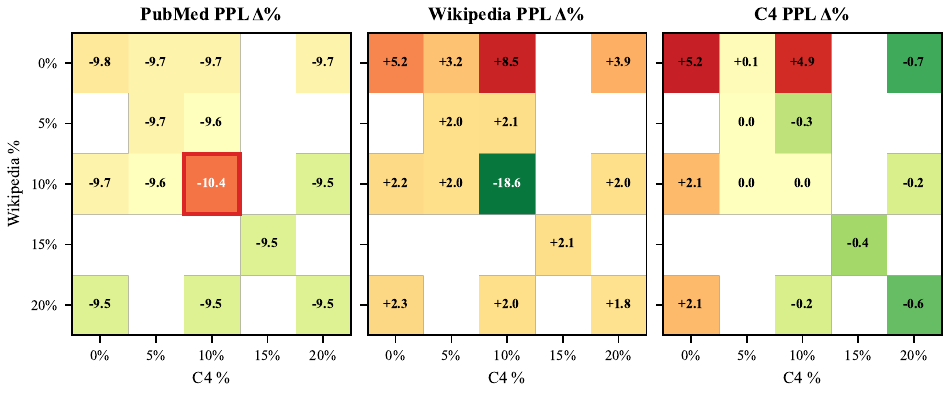}
\caption{Data-mixture ablation heatmap showing the perplexity change (\%) relative to the base model for PubMed, Wikipedia, and C4. Each cell corresponds to a C4/Wikipedia mixture ratio; the red box highlights the selected 10\%/10\% configuration.}
\label{fig:mixture_heatmap}
\end{figure}

\paragraph{Continued pretraining ablation before supervised fine-tuning.}
The second ablation examined the role of continued pretraining before supervised fine-tuning. At 130M (Table~\ref{tab:sft_biomamba_only}), continued pretraining increased BioASQ accuracy from 53.66\% (44/82; 95\% confidence interval, 42.9\% to 64.0\%) to 68.29\% (56/82; 95\% confidence interval, 57.6\% to 77.4\%), with macro-F1 rising from 0.536 to 0.634; PubMedQA accuracy increased from 56.00\% (112/200; 95\% confidence interval, 49.1\% to 62.7\%) to 63.00\% (126/200; 95\% confidence interval, 56.1\% to 69.4\%), with macro-F1 from 0.570 to 0.661. The class-specific pattern differs between benchmarks: on BioASQ the yes F1 rose from 0.513 to 0.768 while the no F1 dropped from 0.558 to 0.500, indicating that in the smallest model biomedical pretraining mainly strengthened recognition of affirmative evidence patterns; on PubMedQA both yes and no F1 improved (0.667$\to$0.727 and 0.472$\to$0.595), consistent with a more symmetric benefit (Supplementary Table~\ref{tab:sft_bioasq_detailed}).

Across scales (Table~\ref{tab:sft_biomamba_only}), on PubMedQA continued pretraining improved accuracy at every scale, with gains ranging from 0.50 to 7.00 percentage points. On BioASQ, the gain was large at 130M, modest at 780M and 1.3B, neutral at 2.7B, and slightly negative at 370M. These results show that the value of biomedical continued pretraining after task-specific supervision depends on both model scale and downstream task.

\subsection{Learning curves and convergence}

\begin{figure}[!htbp]
\centering
\begin{subfigure}{0.48\linewidth}
\centering
\safeincludegraphics[width=\linewidth]{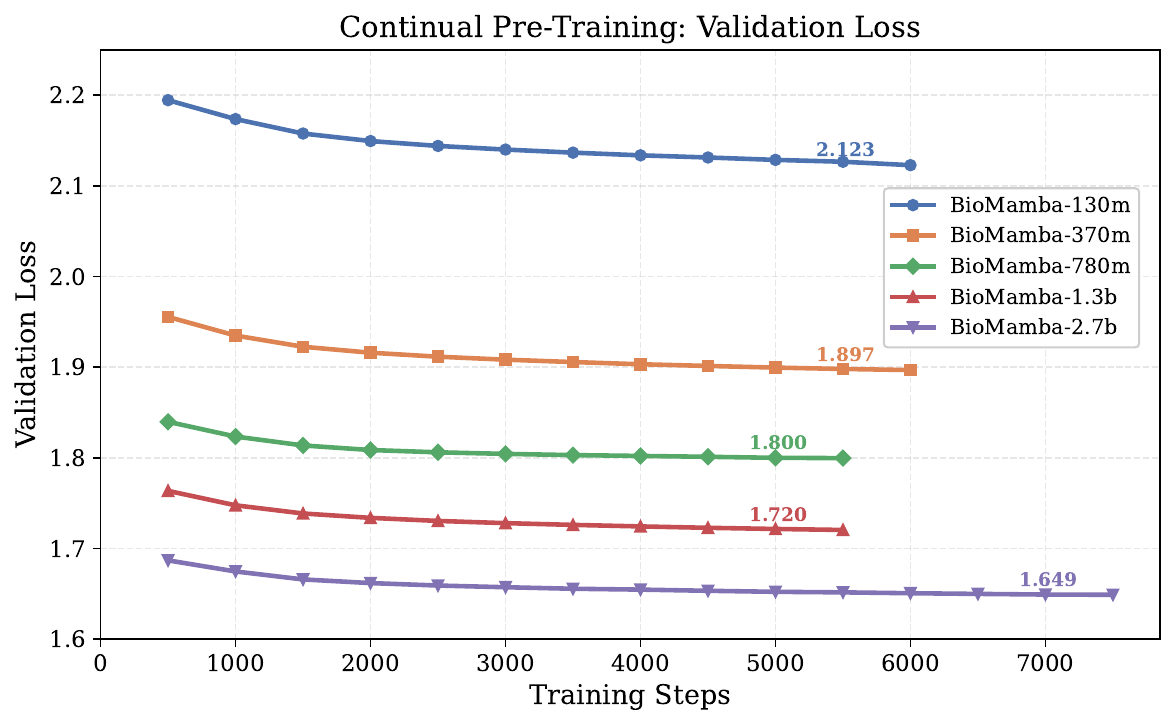}
\caption{Validation loss.}
\end{subfigure}
\hfill
\begin{subfigure}{0.48\linewidth}
\centering
\safeincludegraphics[width=\linewidth]{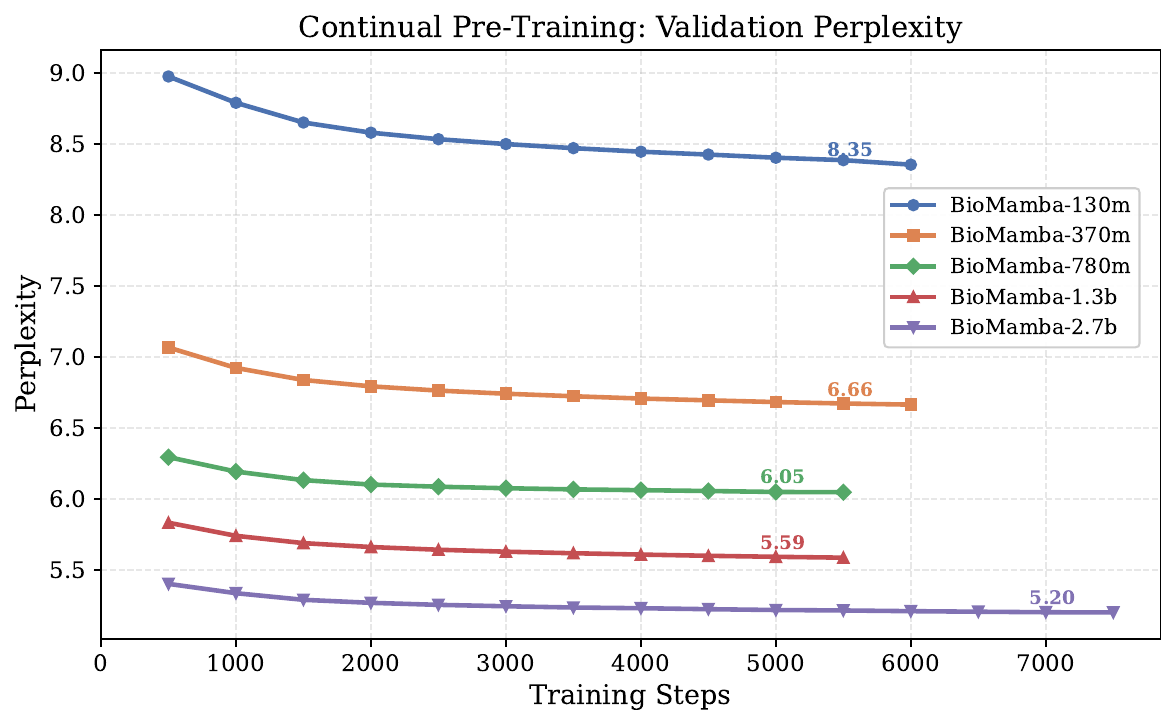}
\caption{Validation perplexity.}
\end{subfigure}
\caption{Training dynamics during continued pretraining. All model scales converged smoothly, and the best validation point occurred at the final step of every run. The 3-epoch continued pretraining budget was chosen because validation loss had effectively plateaued by the end of training; additional training epochs produced diminishing reductions in validation loss, so the 3-epoch schedule is convergence-driven rather than a fixed-budget compromise.}
\label{fig:cpt_dynamics}
\end{figure}

Optimization remained stable across all scales (Figure~\ref{fig:cpt_dynamics}): validation loss and perplexity decreased smoothly during continued pretraining, and the best validation point occurred at the final step in every run (lowest validation losses: 2.11 / 1.90 / 1.80 / 1.72 / 1.65 at 130M / 370M / 780M / 1.3B / 2.7B). These patterns indicate that the three-epoch schedule was well calibrated and that the reported language-model improvements are unlikely to be artifacts of unstable optimization or selective early stopping.

\section{Discussion}
\label{sec:discussion}

This study examined whether biomedical continued pretraining can improve Mamba language models without substantially weakening general-language-modeling fluency. The results support this goal across both biomedical literature and clinical text. Under tokenizer-controlled internal evaluation, BioMamba lowered PubMed perplexity at every model scale, improved Wikipedia-style held-out perplexity, and left C4 performance largely unchanged; a six-task out-of-domain multiple-choice suite and an evaluation-budget stability check (Section~\ref{sec:ood_longctx}) further support preservation on the benchmarks tested. After supervised fine-tuning, the models also transferred across three downstream tasks: note completion, discharge summary generation, and biomedical yes/no question answering. Taken together, the findings support balanced domain adaptation as a practical strategy for biomedical state space language models.

\subsection{Practical biomedical and clinical applications}

The most immediate use of BioMamba is workflow support rather than autonomous bedside decision making. The PubMed, BioASQ, and PubMedQA results are most relevant to tasks such as literature screening, evidence extraction, abstract triage, biomedical question answering, and drafting structured evidence summaries~\cite{yue2024clinicalagent,yue2024trialenroll}. The MIMIC-IV results extend this picture to clinical documentation support, especially note completion and discharge summary generation, which are closely related to hospital-course summarization and discharge-note generation tasks studied previously \cite{Adams2021WhatsIA,Hartman2022ADA,Searle2022DischargeSH,Aali2024ADA,Small2025EvaluatingHC}. These tasks require models to handle biomedical terminology together with broader scientific or clinical language, which fits the balanced adaptation strategy used here. Recent studies of medical large language models report strong performance on clinical knowledge and medical question answering benchmarks \cite{Peng2023ASO,Singhal2022LargeLM,Singhal2025TowardEM,yang2025ecg}, but discussions of clinical deployment similarly suggest that the most credible near-term role of such models is workflow support with retrieval and human review rather than autonomous clinical decision making \cite{artsi2025large,liu2025rag_biomedicine}.

The scale analysis also has practical implications. Continued pretraining produced its largest downstream gains in the smaller models, suggesting that useful biomedical adaptation is not limited to the largest checkpoint. This is relevant for institutions that require lower inference cost, local deployment, or more reproducible fine-tuning under limited computational budgets. At the same time, the present results do not justify direct use for patient-specific diagnosis, prognosis, or treatment recommendation. BioMamba was trained mainly on PubMed-centered text and evaluated on benchmark question answering and clinical generation rather than on prospective clinical tasks. Its most suitable role is therefore as an assistive component in literature-grounded or documentation-support pipelines with retrieval, source attribution, and expert oversight \cite{agrawal2025evaluation_illusion,artsi2025large}.

\subsection{Related Work}

Biomedical language modeling has been dominated by Transformer-based architectures, including SciBERT \cite{beltagy2019scibert}, BioBERT \cite{lee2020biobert}, PubMedBERT \cite{gu2021domain}, BioGPT \cite{luo2022biogpt}, SciFive \cite{Phan2021SciFiveAT}, BioBART \cite{Yuan2022BioBARTPA}, and BioMedLM \cite{bolton2024biomedlm}, which established the value of domain-specific pretraining for biomedical natural language processing. Text-to-text pretraining paradigms such as T5 \cite{Raffel2019ExploringTL} and link-aware literature pretraining approaches such as LinkBERT \cite{Yasunaga2022LinkBERTPL} further expanded the design space for biomedical language modeling. Clinical-domain models such as ClinicalBERT \cite{Alsentzer2019PubliclyAC}, GatorTron \cite{Yang2022GatorTronAL}, ClinicalT5 \cite{Lu2022ClinicalT5AG}, and broader EHR-oriented language models \cite{Yang2022ALL,Kweon2023PubliclySC} extended domain adaptation to clinical notes~\cite{li2025bridging}. Work on hospital-course summarization and clinical text generation has also expanded, including dataset construction, model adaptation, and shared-task evaluation \cite{Adams2021WhatsIA,Hartman2022ADA,Searle2022DischargeSH,Aali2024ADA,Xu2024OverviewOT,Small2025EvaluatingHC}. Biomedical question answering has a mature benchmark ecosystem centered on BioASQ \cite{Tsatsaronis2015AnOO,Krithara2022BioASQQAAM,Nentidis2023BioASQ,nentidis2020bioasq,Nentidis2025OverviewOB} and PubMedQA \cite{jin2019pubmedqa}. More recent open medical language models such as PMC-LLaMA \cite{Wu2023PMCLLaMATB}, BioMistral \cite{Labrak2024BioMistralAC}, MEDITRON-70B \cite{Chen2023MEDITRON70BSM}, and Clinical Camel \cite{Toma2023ClinicalCA} still rely mainly on Transformer-style backbones. In parallel, structured state space models \cite{Gu2021EfficientlyML}, Mamba \cite{gu2023mamba}, and state space duality refinements \cite{dao2024transformers} have provided an alternative framework with favorable scaling for long inputs~\cite{xu2024smiles}. However, biomedical adaptation of these models remains limited, and published work has so far focused more narrowly on settings such as clinical text modeling \cite{yang2024clinicalmamba}.

The present study differs from much of this literature in two ways. First, the main evidence is based on tokenizer-controlled comparisons between each released base checkpoint and the corresponding checkpoint after continued pretraining. This design isolates the effect of biomedical adaptation more directly than comparisons across unrelated public models. Second, the adaptation strategy explicitly addresses the trade-off between biomedical specialization and retention of general-language-modeling fluency. That trade-off matters in biomedical literature workflows, where specialized terminology is often mixed with broader scientific language, and it has been emphasized in continued pretraining studies \cite{Gururangan2020DontSP,Ke2023AdaptingAL,Parmar2024ReuseDR}. The question also relates to ongoing debate about whether generalist foundation models can replace specialized biomedical or clinical models \cite{Lehman2023DoWS,Nori2023CanGF}. Recent benchmark studies in biomedical NLP and real-world clinical practice text \cite{Chen2023BenchmarkingLL,Wu2025BRIDGEBL}, together with cautionary few-shot results in the biomedical domain \cite{Moradi2021GPT3MA}, suggest that conclusions depend strongly on task formulation and evaluation protocol. Recent methodological work has also argued that benchmark gains should be interpreted cautiously and connected to realistic biomedical use settings \cite{agrawal2025evaluation_illusion}. In that context, the present study aims to provide controlled and reproducible evidence on biomedical continued pretraining for Mamba-family models rather than to claim clinical readiness.

\subsection{New Findings}

Several findings merit emphasis.

First, the mixture ratio mattered more than the raw inclusion of non-biomedical data. At 130M, PubMed-only continued pretraining worsened Wikipedia-style held-out perplexity by 5.2\% and C4 perplexity by 5.2\% relative to the base model, and most alternative mixtures we tested retained the PubMed gain but still weakened at least one held-out benchmark. At 370M, the PubMed-only control was worse than the 80/10/10 recipe on Wikipedia (+69\%), C4 (+15\%), and even held-out PubMed itself (9.32 $\to$ 9.91; Supplementary Table~\ref{tab:mixing_370m}), indicating that the forgetting penalty grows with scale and that adding general-domain data did not trade off against in-domain performance. The 80/10/10 recipe therefore appears effective because it strikes a specific balance rather than because it simply adds more data, and this balance is load-bearing beyond the original 130M pilot.

Second, the downstream effect of continued pretraining after supervised fine-tuning was jointly task- and scale-dependent: PubMedQA gains were positive at every scale, MIMIC-IV gains were positive at every scale on both completion and discharge generation, while BioASQ gains were non-monotonic and the 370M and 2.7B deltas fell within the 1.22 pp single-question resolution and the seed-level variance reported in Supplementary Table~\ref{tab:multiseed_qa}. A likely explanation is that smaller models benefit more from additional domain exposure, whereas larger models may already contain enough transferable knowledge for some supervised tasks---yet biomedical adaptation and scaling remained complementary, since the BioMamba+SFT advantage over Mamba2+SFT persisted at every evaluated scale even where absolute ROUGE plateaued at 1.3B on MIMIC-IV.

\subsection{Limitations}

Several limitations should be considered when interpreting these results. The continued pretraining corpus was restricted to PubMed abstracts and did not include full-text biomedical articles, clinical narratives, electronic health records, or multimodal clinical data~\cite{ren2024moving}. The downstream evaluation, although broader than yes/no question answering alone, remained limited to three task families---note completion, discharge summary generation, and yes/no question answering---and therefore does not cover the full range of biomedical language model utility. Additional tasks such as information extraction, retrieval, evidence attribution, broader summarization, and long-context reasoning were not evaluated, despite their prominence in recent biomedical and clinical benchmark suites \cite{Chen2023BenchmarkingLL,Wu2025BRIDGEBL}.

The mixture-ratio search was performed primarily in a 130M pilot study, and the selected mixture was then applied across all model scales, with a single 370M cross-scale ablation confirming that the mixture is load-bearing at larger scales. Although the cross-scale results were broadly consistent, this design does not exclude the possibility that different scales would benefit from different data proportions or training schedules.

Our comparisons with publicly released biomedical checkpoints (Tables~\ref{tab:external_comparison} and \ref{tab:mimic_external}) are contextual rather than controlled, as discussed in the Methods and Results sections, and should not be read as leaderboard-style performance claims. In addition, the benchmark datasets were modest in size, especially the BioASQ yes/no evaluation set, so small absolute differences should be interpreted cautiously.

A further limitation is that the study did not assess calibration, uncertainty estimation, hallucination behavior, robustness to distribution shift, or safety in real clinical workflows~\cite{wan2025evaluating}. Our factuality analysis (Section~\ref{sec:results}) uses NLI contradiction as a proxy for source faithfulness; this does not replace a clinician audit, and the current study does not establish clinical safety, uncertainty calibration, or workflow readiness. A prospective clinician audit with a structured error taxonomy is left to future work. The evaluation scope of this study is in-domain biomedical QA and clinical generation, together with general-language-modeling fluency; broader capabilities such as retrieval-grounded reasoning, evidence-grounded citation, structured information extraction, long-context clinical reasoning, and prospective deployment were not evaluated and are therefore hypothesized rather than demonstrated. Improvement on benchmark question answering or clinical generation does not by itself establish clinical validity or deployment readiness \cite{agrawal2025evaluation_illusion}. The present results should therefore be interpreted as evidence about language-model adaptation and transfer, not as evidence that BioMamba is suitable for autonomous clinical use.

\subsection{Future Work}

Several directions follow directly from these limitations. An immediate next step is to expand the continued pretraining corpus to include full-text biomedical articles, clinical guidelines, trial reports, and carefully governed clinical narratives when available. It would also be useful to study whether the optimal PubMed/C4/Wikipedia balance changes with model size, training budget, or target task. More systematic ablations on data mixture, training duration, and learning-rate schedules may further refine the adaptation strategy.

Future evaluation should extend beyond the present tasks of note completion, discharge summary generation, and yes/no question answering to tasks that are closer to biomedical and clinical workflows~\cite{agnikula2021social}, including retrieval-augmented question answering, evidence extraction, literature screening, broader summarization, uncertainty estimation, and source-grounded generation \cite{liu2025rag_biomedicine,artsi2025large}. It will also be important to assess robustness, calibration, and error patterns in settings with expert review. More broadly, the present results suggest that biomedical Mamba models may be most useful as components in retrieval-grounded and clinician-supervised systems rather than as standalone models.

\section{Conclusion}

BioMamba shows that biomedical domain-adaptive continued pretraining can strengthen public Mamba2 checkpoints without changing the underlying architecture. Across five model scales from 130M to 2.7B, BioMamba improves biomedical language modeling, preserves general-language-modeling fluency through a balanced 80\%/10\%/10\% PubMed/C4/Wikipedia training corpus, and transfers effectively to three downstream tasks: clinical note completion, discharge summary generation, and biomedical question answering. Beyond the empirical findings, our main contribution is a reproducible biomedical continued pretraining recipe together with a family of open-weight BioMamba checkpoints (130M / 370M / 780M / 1.3B / 2.7B), released publicly with the full training and evaluation code to provide the biomedical natural language processing community with a reusable resource.

\section*{Ethical Approval}
No new human-subject data were collected. This study used public text corpora and benchmark datasets together with de-identified clinical notes from MIMIC-IV-Note v2.2, accessed under PhysioNet's credentialed data use terms. All figures in this manuscript were created by the authors, who hold the copyright.

\section*{Data Availability}
Training and evaluation code, configurations, and the five BioMamba checkpoints (130M / 370M / 780M / 1.3B / 2.7B) are publicly available at \url{https://github.com/lingyue404/BioMamba}; the released checkpoints are also collected at \url{https://huggingface.co/collections/zmzfpc/biomamba}. The release is under the Apache-2.0 license, matching the upstream Mamba2 checkpoints. The continued pretraining corpora were loaded from Hugging Face as \texttt{cyrilzakka/pubmed-medline} (\texttt{default}, revision \texttt{432681e19469e93e6c42878d5f41fec400974fb8}), \texttt{wikimedia/wikipedia} (\texttt{20231101.en}, revision \texttt{b04c8d1ceb2f5cd4588862100d08de323dccfbaa}), and \texttt{allenai/c4} (\texttt{en}, revision \texttt{1588ec454efa1a09f29cd18ddd04fe05fc8653a2}); the Wikipedia-style held-out evaluation set uses the Wikitext-103 raw split, as described in Supplementary Section~\ref{app:reproducibility}. BioASQ and PubMedQA are publicly available from their original sources. MIMIC-IV-Note v2.2 is available to credentialed users through PhysioNet under its data-use requirements. Public checkpoint identifiers used for the contextual comparisons are listed in Supplementary Section~\ref{app:eval_details}, and additional reproducibility details (corpus snapshots, held-out construction, preprocessing rules, prompt templates, and model-selection criteria) are provided in Supplementary Section~\ref{app:reproducibility}.

\section*{Funding}
Tianfan Fu is supported by Young Scientists Fund (C Class) of the National Natural Science Foundation of China (Grant No.~62506154), the Fundamental Research Funds for the Central Universities and Nanjing University International Collaboration Initiative (Grant No.~020214380129) and the “111 Center”(No.~B26023). Yanbo Wang is supported by the Joint Funds of the National Natural Science Foundation of China (No. U21A20524).

\section*{Supplementary Materials}
All supplementary materials are provided as an appendix to this manuscript.

\section*{Acknowledgments} 

Tianfan Fu is supported by Young Scientists Fund (C Class) of the National Natural Science Foundation of China (Grant No. 62506154), the Fundamental Research Funds for the Central Universities and Nanjing University International Collaboration Initiative (Grant No. 020214380129) and the “111 Center” (No. B26023). 
Yanbo Wang is supported by the Joint Funds of the National Natural Science Foundation of China (No. U21A20524). 
We thank the BioASQ organizers, the PubMedQA authors, and the broader biomedical natural language processing community for maintaining the open benchmarks and open-source tooling that made these experiments possible. We also thank Liantao Ma (Peking University) and Yinghao Zhu (The University of Hong Kong) for assistance with experiments and data processing.

\printbibliography

\appendix
\renewcommand{\thesection}{S\arabic{section}}
\renewcommand{\thetable}{S\arabic{table}}
\renewcommand{\thefigure}{S\arabic{figure}}
\setcounter{section}{0}
\setcounter{table}{0}
\setcounter{figure}{0}

\section*{Supplementary Materials}

This supplement provides implementation details, evaluation details, exact public checkpoint sources, class-specific downstream results, additional clinical generation case studies, and supplementary tables covering the cross-scale mixture ablation, out-of-domain evaluation and evaluation-budget stability, matched-budget SFT control, 3-seed QA reproduction, BioASQ item-level error analysis, NLI factuality protocol, MIMIC output diagnostics, bootstrap confidence intervals, and reproducibility details.

\section{Experimental Environment}
\label{app:environment}

All continued pretraining and language-model evaluation experiments were conducted in the environment summarized in Table~\ref{tab:env}.

\begin{table}[htbp]
\centering
\caption{Experimental environment for continued pretraining and evaluation.}
\label{tab:env}
\vspace{1mm}
\begin{tabular}{@{}L{0.5\textwidth}L{0.43\textwidth}@{}}
\toprule
\textbf{Item} & \textbf{Version / Specification} \\
\midrule
GPU & NVIDIA H100 80 GB HBM3 \\
GPU driver & 580.82.07 \\
CUDA & 12.8 \\
Operating system & Ubuntu 24.04.1 LTS (Kernel 5.14.0) \\
Python & 3.12.3 \\
PyTorch & 2.10.0+cu128 \\
\texttt{mamba-ssm} & 2.3.0 \\
\texttt{transformers} & 4.57.1 \\
\texttt{datasets} & 4.0.0 \\
\bottomrule
\end{tabular}
\end{table}

\section{Continued Pretraining Implementation Details}
\label{app:cpt_impl}

Table~\ref{tab:hyperparams} summarizes the hyperparameters shared across BioMamba continued pretraining runs, and Table~\ref{tab:per_size} provides per-scale implementation details.

\begin{table}[htbp]
\centering
\caption{Shared continued pretraining hyperparameters.}
\label{tab:hyperparams}
\vspace{1mm}
\begin{tabular}{@{}L{0.34\linewidth}L{0.14\linewidth}L{0.46\linewidth}@{}}
\toprule
\textbf{Parameter} & \textbf{Value} & \textbf{Description} \\
\midrule
Epochs & 3 & Training epochs \\
Weight decay & 0.1 & AdamW weight decay \\
Maximum sequence length & 1,024 & Tokens per sequence \\
Gradient clipping & 1.0 & Maximum gradient norm \\
Precision & bfloat16 & Mixed precision training \\
Seed & 42 & Random seed \\
\midrule
\multicolumn{3}{@{}l@{}}{\textit{Warmup-stable-decay scheduler}} \\
Warmup ratio & 0.10 & Linear warmup \\
Stable ratio & 0.65 & Constant learning-rate plateau \\
Decay ratio & 0.20 & Cosine decay \\
Minimum learning-rate ratio & 0.1 & Minimum learning-rate ratio \\
\midrule
\multicolumn{3}{@{}l@{}}{\textit{Advanced techniques}} \\
Exponential moving average & Enabled & Decay = 0.999 \\
Layer-wise learning-rate decay & 0.90--0.95 & Lower layers adapt more conservatively \\
\bottomrule
\end{tabular}
\end{table}

\begin{table}[htbp]
\centering
\caption{Per-scale continued pretraining configuration, combining optimization (batch, gradient accumulation, effective batch size, layer-wise learning-rate decay, maximum learning rate, total run steps) and resource usage (number of GPUs, peak memory per GPU). The base learning rate was set conservatively and scaled by model size to reduce catastrophic forgetting during continued pretraining.}
\label{tab:per_size}
\footnotesize
\setlength{\tabcolsep}{3pt}
\begin{tabular}{@{}C{0.06\linewidth}C{0.11\linewidth}C{0.08\linewidth}C{0.09\linewidth}C{0.10\linewidth}C{0.10\linewidth}C{0.08\linewidth}C{0.10\linewidth}C{0.12\linewidth}@{}}
\toprule
\textbf{Scale} & \textbf{Batch/GPU} & \textbf{Accum} & \textbf{Eff. BS} & \textbf{LR decay} & \textbf{Max. LR} & \textbf{Steps} & \textbf{\# GPUs} & \textbf{Peak mem.} \\
\midrule
130M & 32 & 8  & 256 & 0.90 & 1.5e-6 & 5,967 & 1$\times$ H100 & $\sim$40 GB \\
370M & 48 & 5  & 240 & 0.95 & 4.5e-7 & 6,363 & 1$\times$ H100 & $\sim$76 GB \\
780M & 32 & 8  & 256 & 0.95 & 3.0e-7 & 5,967 & 1$\times$ H100 & $\sim$65 GB \\
1.3B & 16 & 16 & 256 & 0.95 & 2.5e-7 & 5,967 & 1$\times$ H100 & $\sim$65 GB \\
2.7B & 6  & 8  & 192 & 0.95 & 2.0e-7 & 7,953 & 4$\times$ H100 & $\sim$80 GB \\
\bottomrule
\end{tabular}
\end{table}

\section{Evaluation Details and Public Checkpoint Sources}
\label{app:eval_details}

The exact public checkpoints used for the contextual comparison in the main text are listed in Table~\ref{tab:external_sources}. Internal evaluations followed the fixed PubMed, Wikipedia, and C4 validation sets described in the main Methods section, whereas public-checkpoint comparisons used the same raw text with each model's native tokenizer.

\paragraph{Loss computation.}
Perplexity was computed from next-token cross-entropy with padding masked via \texttt{ignore\_index} = $-100$. Non-finite losses were excluded from averaging.

\begin{table}[htbp]
\centering
\caption{Exact public checkpoint sources used for the contextual comparison in the main text.}
\label{tab:external_sources}
\vspace{1mm}
\small
\begin{tabular}{@{}L{0.26\textwidth}L{0.64\textwidth}@{}}
\toprule
\textbf{Model} & \textbf{Public checkpoint} \\
\midrule
BioGPT & \texttt{microsoft/biogpt} \\
BioGPT-Large & \texttt{microsoft/BioGPT-Large} \\
BioGPT-Large-PubMedQA & \texttt{microsoft/BioGPT-Large-PubMedQA} \\
BioMedLM & \texttt{stanford-crfm/BioMedLM} \\
Meditron3-Gemma2-2B & \texttt{OpenMeditron/Meditron3-Gemma2-2B} \\
Bio-Medical-Llama-3.2-1B & \path{ContactDoctor/Bio-Medical-Llama-3-2-1B-CoT-012025} \\
Gemma3-1B (fine-tuned) & \texttt{kunjcr2/gemma3\_finetune} \\
\bottomrule
\end{tabular}
\end{table}

\section{Supervised Fine-Tuning Details}
\label{app:sft_details}

Table~\ref{tab:sft_config} summarizes the selected supervised fine-tuning hyperparameters for each BioMamba scale. Table~\ref{tab:sft_bioasq_detailed} provides class-specific yes/no F1 breakdowns for BioASQ and PubMedQA across scales.

\begin{table}[htbp]
\centering
\caption{Supervised fine-tuning hyperparameters. ``BioASQ share'' denotes the proportion of BioASQ data in the training mixture.}
\label{tab:sft_config}
\begin{tabular}{@{}L{0.15\linewidth}C{0.15\linewidth}C{0.11\linewidth}C{0.17\linewidth}C{0.13\linewidth}C{0.13\linewidth}@{}}
\toprule
\textbf{Model} & \textbf{Learning rate} & \textbf{Epochs} & \textbf{Effective batch size} & \textbf{Weight decay} & \textbf{BioASQ share} \\
\midrule
130M & 3e-5 & 3 & 128 & 0.01 & 0.65 \\
370M & 3e-5 & 2 & 64 & 0.01 & 0.70 \\
780M & 2e-5 & 5 & 64 & 0.03 & 0.65 \\
1.3B & 7e-6 & 3 & 32 & 0.03 & 0.70 \\
2.7B & 2e-5 & 5 & 32 & 0.01 & 0.65 \\
\bottomrule
\end{tabular}
\end{table}

\begin{table}[htbp]
\centering
\caption{Class-specific yes/no F1 for BioASQ and PubMedQA across model scales. M2+SFT = Mamba2+SFT (the public Mamba2 base checkpoint after supervised fine-tuning); BM+SFT = BioMamba+SFT (the counterpart after continued pretraining and the same supervised fine-tuning). Accuracy and macro-F1 are reported in the main text (Table~\ref{tab:sft_biomamba_only}); this table shows only the yes/no F1 breakdown to avoid repetition.}
\label{tab:sft_bioasq_detailed}
\setlength{\tabcolsep}{5pt}
\begin{tabular}{@{}L{0.18\linewidth}C{0.10\linewidth}L{0.12\linewidth}C{0.12\linewidth}C{0.12\linewidth}@{}}
\toprule
\textbf{Benchmark} & \textbf{Scale} & \textbf{Arm} & \textbf{Yes F1} & \textbf{No F1} \\
\midrule
\multirow{10}{*}{BioASQ 13B} & \multirow{2}{*}{130M} & M2+SFT & 0.513 & 0.558 \\
 & & BM+SFT & \textbf{0.768} & 0.500 \\
\cmidrule(l){2-5}
 & \multirow{2}{*}{370M} & M2+SFT & \textbf{0.862} & 0.585 \\
 & & BM+SFT & 0.850 & \textbf{0.591} \\
\cmidrule(l){2-5}
 & \multirow{2}{*}{780M} & M2+SFT & \textbf{0.847} & 0.609 \\
 & & BM+SFT & 0.835 & \textbf{0.721} \\
\cmidrule(l){2-5}
 & \multirow{2}{*}{1.3B} & M2+SFT & 0.877 & 0.720 \\
 & & BM+SFT & \textbf{0.893} & \textbf{0.769} \\
\cmidrule(l){2-5}
 & \multirow{2}{*}{2.7B} & M2+SFT & \textbf{0.927} & \textbf{0.852} \\
 & & BM+SFT & \textbf{0.927} & \textbf{0.852} \\
\midrule
\multirow{10}{*}{PubMedQA} & \multirow{2}{*}{130M} & M2+SFT & 0.667 & 0.472 \\
 & & BM+SFT & \textbf{0.727} & \textbf{0.595} \\
\cmidrule(l){2-5}
 & \multirow{2}{*}{370M} & M2+SFT & 0.736 & \textbf{0.639} \\
 & & BM+SFT & \textbf{0.746} & 0.633 \\
\cmidrule(l){2-5}
 & \multirow{2}{*}{780M} & M2+SFT & 0.737 & 0.636 \\
 & & BM+SFT & \textbf{0.754} & \textbf{0.667} \\
\cmidrule(l){2-5}
 & \multirow{2}{*}{1.3B} & M2+SFT & 0.737 & 0.619 \\
 & & BM+SFT & \textbf{0.782} & \textbf{0.697} \\
\cmidrule(l){2-5}
 & \multirow{2}{*}{2.7B} & M2+SFT & 0.783 & 0.703 \\
 & & BM+SFT & \textbf{0.800} & \textbf{0.745} \\
\bottomrule
\end{tabular}
\end{table}

\section{MIMIC-IV Clinical SFT and Evaluation Details}
\label{app:mimic_details}

This section provides full details for the MIMIC-IV clinical generation experiments reported in the main text.

\subsection{Data source and splits}

The evaluation corpus is MIMIC-IV-Note v2.2 \cite{PhysioNet-mimic-iv-note-2.2}, derived from the broader MIMIC-IV resource \cite{johnson2023mimiciv}, and contains 331,793 de-identified discharge summaries from 145,914 unique patients (average note length: 10,551 characters). Data were split by \texttt{subject\_id} at a 9:1 ratio (seed = 42), yielding approximately 131,323 training patients (298,614 notes) and 14,591 test patients (33,179 notes).

\subsection{SFT data construction}

Two supervised fine-tuning tasks were constructed from the discharge notes:

\paragraph{Note completion.} The first 50\% of each note serves as the instruction; the remainder is the response. Filtering requires $\geq$100 characters for the prefix and $\geq$50 characters for the continuation (\texttt{max\_context\_chars} = 3{,}000; \texttt{max\_response\_chars} = 2{,}000). Up to 20{,}000 samples were retained.

\paragraph{Discharge summary generation.} Admission sections (chief complaint, history of present illness, past medical history, medications, allergies, social and family history, physical examination, and laboratory results) form the instruction. Discharge sections (hospital course, assessment and plan, discharge diagnosis, discharge medications, discharge instructions, and discharge condition) form the response. The same character limits and filters apply, yielding approximately 12{,}893 valid samples.

The combined dataset contains 32{,}893 samples (90\% train / 10\% validation). On average each sample contains 1{,}023 total tokens, of which 139 (13.6\%) are supervised response tokens. Instruction masking sets the loss label of instruction tokens to $-100$.

\subsection{SFT hyperparameters}

Table~\ref{tab:mimic_sft_base} reports the Mamba2+SFT and BioMamba+SFT configurations at 130M, 370M, 780M, and 1.3B after hyperparameter tuning. The 2.7B configurations for both arms were selected by a small sweep over learning rates and epochs following the same conservative pattern; the exact selected recipes and training logs are in the public code release (Supplementary Section~\ref{app:reproducibility}).

\begin{table}[htbp]
\centering
\caption{MIMIC-IV clinical SFT configuration for both arms at 130M--1.3B. Mamba2+SFT uses the baseline recipe; BioMamba+SFT uses lower learning rates and more epochs to preserve domain knowledge acquired during continued pretraining. All runs use AdamW with weight decay 0.01, cosine learning-rate schedule, bf16 precision, gradient clipping 1.0, and early stopping with patience 5; Mamba2+SFT uses 10\% warmup throughout. The 2.7B recipe for both arms is in the public code release (Supplementary Section~\ref{app:reproducibility}).}
\label{tab:mimic_sft_base}
\setlength{\tabcolsep}{3pt}
\begin{tabular}{@{}L{0.17\linewidth}C{0.07\linewidth}C{0.12\linewidth}C{0.14\linewidth}C{0.13\linewidth}C{0.14\linewidth}C{0.08\linewidth}C{0.08\linewidth}@{}}
\toprule
\textbf{Recipe} & \textbf{Size} & \textbf{Batch/GPU} & \textbf{Grad Accum} & \textbf{Effective BS} & \textbf{Learning Rate} & \textbf{Epochs} & \textbf{Warmup} \\
\midrule
\multirow{4}{*}{Mamba2+SFT}   & 130M & 8  & 4  & 32 & 2e-5   & 3 & 0.10 \\
                              & 370M & 4  & 8  & 32 & 2e-5   & 3 & 0.10 \\
                              & 780M & 4  & 8  & 32 & 2e-5   & 3 & 0.10 \\
                              & 1.3B & 2  & 16 & 32 & 2e-5   & 3 & 0.10 \\
\midrule
\multirow{4}{*}{BioMamba+SFT} & 130M & 8  & 4  & 32 & 1e-5   & 5 & 0.10 \\
                              & 370M & 16 & 2  & 32 & 1e-5   & 5 & 0.10 \\
                              & 780M & 16 & 2  & 32 & 7.5e-6 & 6 & 0.12 \\
                              & 1.3B & 16 & 2  & 32 & 9e-6   & 6 & 0.12 \\
\bottomrule
\end{tabular}
\end{table}

\subsection{Evaluation protocol}

Evaluation used 500 randomly sampled test notes per task from the patient-level test split. Both tasks generated up to 128 tokens with greedy decoding (temperature = 0). The maximum context length was 4{,}000 characters. ROUGE scores were computed using the \texttt{rouge-score} library with stemming enabled.

\subsection{BioMamba+SFT tuning observations}

A key finding during hyperparameter tuning is that validation loss does not reliably predict ROUGE quality for BioMamba+SFT models. Lower learning rates often produce higher validation loss but better ROUGE scores. Different model sizes require different learning rates: smaller models (130M--370M) perform best with 1e-5, whereas the 780M and 1.3B models require rates between 7.5e-6 and 9e-6. This suggests that BioMamba models have already internalized domain knowledge and are more sensitive to learning-rate magnitude during clinical fine-tuning.

\section{Additional MIMIC-IV Clinical Generation Case Studies}
\label{app:mimic_case_study}

To complement the representative main-text case studies, we present four additional qualitative examples from the 1.3B model comparing three training stages: Mamba2 (pretrained only), Mamba2+SFT (supervised fine-tuning without continued pretraining), and BioMamba+SFT (continued pretraining followed by supervised fine-tuning). All examples use greedy decoding with a maximum of 128 new tokens.

\subsection{Additional note completion examples}

In the note completion task, the model receives the first 50\% of a discharge note and generates a continuation.

\paragraph{Laboratory results continuation.}

The input ends with a partial complete blood count. Table~\ref{tab:case2} shows that the base model degenerates into repetitive entries, Mamba2+SFT produces correct but incomplete results, and BioMamba+SFT generates three distinct lab panels (basic metabolic panel, minerals, and liver function tests).

\begin{table}[htbp]
\centering
\caption{Additional note-completion case: laboratory results continuation. Input ends with ``\ldots WBC-6.2 RBC-3.23* Hgb-9.7* Hct-30.6* MCV-95 MCH-30.1''.}
\label{tab:case2}
\small
\setlength{\tabcolsep}{4pt}
\begin{tabularx}{\linewidth}{@{}L{0.19\linewidth}Y C{0.08\linewidth}@{}}
\toprule
\textbf{Source} & \textbf{Text} & \textbf{R-1} \\
\midrule
Reference & MCHC-31.8 RDW-15.2 Plt \_\_\_ $\to$ Glucose-70 UreaN-40* Creat-2.4* Na-145 K-4.5 Cl-115* HCO3-21* $\to$ Brief Hospital Course: \_\_\_ year old woman with IDDM, HTN, CKD stage IV\ldots & --- \\
\midrule
Mamba2 & MCHC-32.2* RDW-0.9*\ldots\ then degenerates into repetitive meaningless entries (``BUN-14.1*'' ``BIL-0.9*'' repeated). & 3.96 \\
\midrule
Mamba2+SFT & MCHC-31.8, generates WBC differential (Neuts/Lymphs/Monos), Glucose-137* UreaN-25* Creat-1.9* --- correct format but missing electrolytes. & 5.17 \\
\midrule
\textbf{BioMamba+SFT} & MCHC-31.0, then generates \textbf{complete BMP} (Glucose-137* UreaN-24* Creat-1.6* Na-141 K-4.2 Cl-102 HCO3-26), \textbf{minerals} (Calcium-8.8 Phos-3.5 Mg-1.9), and \textbf{liver function} (ALT-16 AST-20 AlkPhos-76). & \textbf{8.58} \\
\bottomrule
\end{tabularx}
\end{table}

\paragraph{Post-operative course continuation.}

The input describes post-operative pain management. Table~\ref{tab:case3} shows that BioMamba+SFT exactly reproduces the standardized preadmission medication statement used in MIMIC notes and generates a well-structured medication list, achieving ROUGE-1 of 16.43 (+25\% over Mamba2+SFT).

\begin{table}[htbp]
\centering
\caption{Additional note-completion case: post-operative course continuation. Input ends with ``\ldots His pain was well controlled on minimal amounts of oxycodone. He took Colace and senna''.}
\label{tab:case3}
\small
\setlength{\tabcolsep}{4pt}
\begin{tabularx}{\linewidth}{@{}L{0.19\linewidth}Y C{0.08\linewidth}C{0.08\linewidth}@{}}
\toprule
\textbf{Source} & \textbf{Text} & \textbf{R-1} & \textbf{R-2} \\
\midrule
Reference & to prevent constipation. Incision was intact without redness or drainage. He was ambulating well. He received teaching for all of his transplant medications. Medications on Admission: 1.\ Atorvastatin 10\,mg PO DAILY 2.\ Ferrous Sulfat\ldots & --- & --- \\
\midrule
Mamba2 & ``On postop day 4, he was discharged home on metoprolol tartrate, dulcolax suppository, and oxycodone'' --- repeats the discharge description twice and gives an incomplete medication list. & 11.47 & 0.93 \\
\midrule
Mamba2+SFT & ``He was discharged home on postop day 4 in stable condition.'' $\to$ generates reasonable admission and discharge medication lists with transplant-appropriate drugs (Cellcept, Solu-medrol). & 13.15 & 4.71 \\
\midrule
\textbf{BioMamba+SFT} & ``He was discharged home on postop day 3 in stable condition.'' $\to$ \textbf{exactly reproduces}: ``Medications on Admission: The Preadmission Medication list is accurate and complete.'' $\to$ generates a structured medication list (Amlodipine, Aspirin, Atorvastatin, Calcitriol, Docusate, Furosemide). & \textbf{16.43} & \textbf{9.95} \\
\bottomrule
\end{tabularx}
\end{table}

\subsection{Additional discharge-summary generation examples}

In the discharge summary generation task, the model receives structured admission sections (chief complaint, history of present illness, physical exam, labs, etc.) and generates discharge content.

\paragraph{Diverticulitis surgery discharge.}

Table~\ref{tab:case5} shows a clear difference in document planning: BioMamba+SFT spontaneously generates a structured Hospital Course narrative, the most clinically central section of a discharge summary, whereas Mamba2+SFT remains closer to laboratory-style continuation.

\begin{table}[htbp]
\centering
\caption{Additional discharge-summary case: diverticulitis surgery discharge. Input: Service: SURGERY; Chief Complaint: Abdominal pain and failure to thrive related to diverticulitis; Procedure: Low anterior resection, mobilization of splenic flexure\ldots}
\label{tab:case5}
\small
\setlength{\tabcolsep}{4pt}
\begin{tabularx}{\linewidth}{@{}L{0.19\linewidth}Y C{0.08\linewidth}C{0.08\linewidth}@{}}
\toprule
\textbf{Source} & \textbf{Text} & \textbf{R-1} & \textbf{R-2} \\
\midrule
Reference & DISCHARGE MEDICATIONS: 1.\ Acetaminophen 325--650\,mg PO Q4H:PRN pain 2.\ Aspirin 81\,mg PO DAILY 3.\ Depakote 250\,mg BID 4.\ Levothyroxine 75\,mcg PO DAILY 5.\ OxycoDONE \_\_\_ mg PO Q4H:PRN pain\ldots & --- & --- \\
\midrule
Mamba2 & Generates repetitive meaningless lab values (``pH-7.4* O2-95\% CO2-38\% Hct-38.0*'' repeated multiple times). & 2.86 & 0.0 \\
\midrule
Mamba2+SFT & Generates lab results in correct format, but lacks discharge medications and Hospital Course narrative. & 8.14 & 0.91 \\
\midrule
\textbf{BioMamba+SFT} & First generates lab results, then produces \textbf{``Patient is a \_\_\_ year old woman with a history of diverticulitis who presented with abdominal pain and failure to thrive\ldots''}, spontaneously creating a Hospital Course narrative that accurately summarizes the admission reason and surgical procedure. & \textbf{21.55} & \textbf{4.35} \\
\bottomrule
\end{tabularx}
\end{table}

\paragraph{Orthopedic discharge instructions.}

Table~\ref{tab:case6} illustrates the repetition problem: Mamba2+SFT falls into a loop repeating ``Please have your INR checked'' six or more times, whereas BioMamba+SFT generates concise, non-repetitive instructions covering four distinct categories.

\begin{table}[htbp]
\centering
\caption{Additional discharge-summary case: orthopedic discharge instructions. Input: Service: ORTHOPAEDICS; Allergies: Penicillins / Sulfa / Aspirin / Iodine / NSAIDs / Codeine / Gluten / Shellfish\ldots\ (extensive allergy list).}
\label{tab:case6}
\small
\setlength{\tabcolsep}{4pt}
\begin{tabularx}{\linewidth}{@{}L{0.19\linewidth}Y C{0.08\linewidth}C{0.08\linewidth}@{}}
\toprule
\textbf{Source} & \textbf{Text} & \textbf{R-1} & \textbf{R-2} \\
\midrule
Reference & HOSPITAL COURSE: \ldots tolerating a regular diet and feeling well\ldots\ afebrile with stable vital signs\ldots\ pain was adequately controlled on an oral regimen\ldots & --- & --- \\
\midrule
Mamba2 & Repeatedly generates a generic template (``Home \_\_\_, dressing changes as instructed, wound checks, staple removal at two weeks''). & 6.50 & 0.0 \\
\midrule
Mamba2+SFT & Generates DVT prophylaxis (coumadin 4 weeks) and INR monitoring, but \textbf{falls into repetition} (``Please have your INR checked on \_\_\_'' repeated 6+ times). & 11.76 & 3.22 \\
\midrule
\textbf{BioMamba+SFT} & ``DVT Prophylaxis: Please take coumadin for 2 weeks.'' $\to$ ``Resume your home medications unless otherwise directed.'' $\to$ ``Follow up with your PCP.'' $\to$ ``Please follow up with your outpatient physical therapy appointment.'' --- \textbf{concise, well-organized discharge instructions}. & \textbf{14.75} & \textbf{5.42} \\
\bottomrule
\end{tabularx}
\end{table}

\subsection{Summary of qualitative findings}

Table~\ref{tab:case_study_summary} summarizes the capability progression across the three training stages observed in the main-text and supplementary case studies.

\begin{table}[htbp]
\centering
\caption{Qualitative capability comparison across training stages based on 1.3B model case studies.}
\label{tab:case_study_summary}
\small
\setlength{\tabcolsep}{4pt}
\begin{tabularx}{\linewidth}{@{}L{0.24\linewidth}Y Y Y@{}}
\toprule
\textbf{Capability} & \textbf{Mamba2} & \textbf{Mamba2+SFT} & \textbf{BioMamba+SFT} \\
\midrule
Clinical terminology accuracy & Frequent misuse (e.g., ``RRR'' to all systems) & Generally correct & Accurate and comprehensive \\
Repetition avoidance & Severe degeneration & Occasional repetition & Minimal repetition \\
Lab result coverage & Disorganized format & Correct format, incomplete coverage & Correct format, multi-panel coverage \\
Discharge document structure & Cannot generate & Generates basic framework & Structured with appropriate content \\
Medication appropriateness & Irrelevant medications & Partially appropriate & Clinically standard regimens \\
Hospital Course narrative & Cannot generate & Fragmented & Coherent clinical narrative \\
\bottomrule
\end{tabularx}
\end{table}

These additional case studies confirm that the quantitative ROUGE improvements reported in the main text reflect meaningful differences in clinical generation quality. BioMamba+SFT generates more comprehensive clinical content (multi-panel lab results, medication reconciliation, Hospital Course narrative), selects medications aligned with standard care patterns, and produces better-structured discharge documents with less repetitive degeneration. These qualitative advantages originate from domain knowledge acquired during continued pretraining on PubMed literature, which is then effectively leveraged through clinical supervised fine-tuning.

\section{Cross-Scale Data-Mixture Ablation at 370M}
\label{app:mixture_370m}

To verify that the selected 80\%/10\%/10\% PubMed/C4/Wikipedia mixture is not specific to the 130M pilot, we additionally compared the paper recipe against a PubMed-only control at the 370M scale under matched optimizer and scheduler settings (Table~\ref{tab:mixing_370m}). The PubMed-only control worsens Wikipedia-style held-out perplexity by 69\% and C4 perplexity by 15\% relative to the 80\%/10\%/10\% recipe, and is also slightly worse on held-out PubMed (9.32 $\to$ 9.91). This pattern is consistent with the 130M pilot, where the 80\%/10\%/10\% mixture already slightly outperformed PubMed-only on PubMed (8.41 vs.\ 8.47; Table~\ref{tab:mixing_full}); the gap is larger at 370M, indicating that domain-only continued pretraining incurs a larger forgetting penalty at scale even on in-domain text. This confirms that the mixture is load-bearing at larger scale and not an artifact of pilot selection.

\begin{table}[htbp]
\centering
\caption{370M cross-scale validation of the 80\%/10\%/10\% continued pretraining mixture against a PubMed-only control. PubMed-only continued pretraining worsens Wikipedia-style held-out perplexity by +69\% and C4 perplexity by +15\% relative to the 80\%/10\%/10\% recipe, and is also slightly worse on held-out PubMed (9.32 $\to$ 9.91), consistent with the 130M pilot pattern (Table~\ref{tab:mixing_full}).}
\label{tab:mixing_370m}
\begin{tabular}{@{}L{0.34\linewidth}C{0.16\linewidth}C{0.18\linewidth}C{0.16\linewidth}@{}}
\toprule
\textbf{370M CPT recipe} & \textbf{C4 PPL ($\downarrow$)} & \textbf{Wikipedia PPL ($\downarrow$)} & \textbf{PubMed PPL ($\downarrow$)} \\
\midrule
80\%/10\%/10\% (paper recipe) & \textbf{18.67} & \textbf{16.96} & \textbf{9.32} \\
PubMed only & 21.45 & 28.72 & 9.91 \\
$\Delta$ (PubMed-only $-$ paper) & $+2.78$ ($+15$\%) & $+11.76$ ($+69$\%) & $+0.59$ \\
\bottomrule
\end{tabular}
\end{table}

\section{Out-of-Domain Evaluation: Protocol and Per-Task Grid}
\label{app:ood_mcq}

The six out-of-domain multiple-choice benchmarks referenced in Table~\ref{tab:ood_mcq_summary} of the main text---LAMBADA, HellaSwag, ARC-Easy, ARC-Challenge, PIQA, and OpenBookQA---were scored by continuation log-likelihood at a fixed evaluation budget, using each benchmark's standard split (LAMBADA, HellaSwag, OpenBookQA: 5{,}000, 5{,}000, 500 test items; ARC-Easy, ARC-Challenge, PIQA: 570, 299, 1{,}838 items respectively). The same evaluation script was used for Mamba2 and BioMamba at every scale. Table~\ref{tab:ood_mcq_full} reports the full per-task accuracy grid (5 scales $\times$ 6 tasks $\times$ 2 arms, 60 cells total) that underlies the summary in Table~\ref{tab:ood_mcq_summary} of the main text. All per-task $\Delta=(\text{BioMamba}-\text{Mamba2})$ lie inside $\pm 3$\,pp at every scale, and no task shows a systematic regression in either direction. The 370M LAMBADA row is noticeably lower than the 130M and 780M rows for \emph{both} arms (Mamba2 18.70\%; BioMamba 19.62\%); this is a known pipeline artifact at that particular 370M checkpoint and affects the two arms identically, so the BioMamba$-$Mamba2 comparison ($+0.92$\,pp) is unaffected. The evaluation script is released with the training and evaluation code (Supplementary Section~\ref{app:reproducibility}).

\begin{table}[htbp]
\centering
\caption{Full per-task OOD accuracy grid (\%) for Mamba2 base and BioMamba across five scales and six benchmarks. Mean is the unweighted average across the six tasks. Boldface marks the arm with the higher accuracy within each (scale, task) cell; a tie is not bolded.}
\label{tab:ood_mcq_full}
\footnotesize
\setlength{\tabcolsep}{3pt}
\begin{tabular}{@{}C{0.07\linewidth}L{0.18\linewidth}C{0.10\linewidth}C{0.11\linewidth}C{0.09\linewidth}C{0.11\linewidth}C{0.08\linewidth}C{0.09\linewidth}C{0.09\linewidth}@{}}
\toprule
\textbf{Scale} & \textbf{Arm} & \textbf{LAMBADA} & \textbf{HellaSwag} & \textbf{ARC-E} & \textbf{ARC-C} & \textbf{PIQA} & \textbf{OBQA} & \textbf{Mean} \\
\midrule
\multirow{2}{*}{130M} & Mamba2-130M    & \textbf{53.18} & \textbf{32.00} & \textbf{42.63} & 27.42 & 64.53 & 25.00 & \textbf{40.79} \\
                     & BioMamba-130M  & 51.08 & 31.68 & 41.93 & \textbf{28.09} & \textbf{64.85} & \textbf{25.80} & 40.57 \\
\midrule
\multirow{2}{*}{370M} & Mamba2-370M    & 18.70 & 38.88 & \textbf{46.67} & 27.09 & 69.21 & \textbf{27.80} & 38.06 \\
                     & BioMamba-370M  & \textbf{19.62} & \textbf{39.16} & 46.14 & \textbf{28.76} & \textbf{69.80} & 26.20 & \textbf{38.28} \\
\midrule
\multirow{2}{*}{780M} & Mamba2-780M    & \textbf{55.52} & 45.80 & 52.98 & \textbf{31.10} & \textbf{72.52} & \textbf{31.60} & \textbf{48.26} \\
                     & BioMamba-780M  & 54.52 & \textbf{46.00} & \textbf{53.51} & 30.10 & 71.93 & 30.40 & 47.74 \\
\midrule
\multirow{2}{*}{1.3B} & Mamba2-1.3B    & 51.22 & 49.34 & 59.65 & \textbf{35.12} & 74.32 & 30.60 & 50.04 \\
                     & BioMamba-1.3B  & \textbf{53.58} & \textbf{49.64} & \textbf{60.35} & 34.78 & \textbf{74.76} & 30.60 & \textbf{50.62} \\
\midrule
\multirow{2}{*}{2.7B} & Mamba2-2.7B    & \textbf{63.86} & 52.82 & 62.98 & \textbf{35.45} & 76.01 & 34.20 & 54.22 \\
                     & BioMamba-2.7B  & 63.82 & \textbf{53.28} & \textbf{63.16} & 34.78 & \textbf{76.06} & \textbf{34.80} & \textbf{54.32} \\
\bottomrule
\end{tabular}
\end{table}

\section{Matched-Budget SFT Control at 1.3B}
\label{app:matched_budget}

To verify that the BioMamba+SFT advantage is not an artifact of the more conservative BioMamba SFT schedule, we additionally fine-tuned Mamba2-1.3B with the exact BioMamba SFT recipe reported in Table~\ref{tab:mimic_sft_base} (lower learning rate, more epochs, same effective batch size). Table~\ref{tab:matched_budget_sft} shows that the BioMamba schedule does not transfer the BioMamba gain to Mamba2; instead, it hurts Mamba2 relative to its own recipe on both completion and discharge generation. This is the expected behavior if BioMamba+SFT is leveraging information acquired during continued pretraining rather than exploiting an unequal optimization effort.

\begin{table}[htbp]
\centering
\caption{Matched-budget 1.3B MIMIC-IV control: Mamba2-1.3B trained with the exact BioMamba SFT recipe (Table~\ref{tab:mimic_sft_base}). The BioMamba schedule worsens Mamba2 relative to its own recipe, which rules out unequal SFT budgets as a source of the BioMamba+SFT advantage. ROUGE \%, 500 test samples, greedy decoding.}
\label{tab:matched_budget_sft}
\small
\setlength{\tabcolsep}{4pt}
\begin{tabular}{@{}L{0.40\linewidth}C{0.25\linewidth}C{0.25\linewidth}@{}}
\toprule
\textbf{1.3B arm (recipe)} & \textbf{Completion R-1 / R-2 / R-L} & \textbf{Discharge R-1 / R-2 / R-L} \\
\midrule
Paper Mamba2+SFT  & 7.93 / 3.28 / 5.76 & 9.99 / 3.70 / 7.04 \\
Paper BioMamba+SFT & \textbf{8.11} / \textbf{3.33} / \textbf{5.89} & \textbf{10.11} / 3.69 / \textbf{7.04} \\
\midrule
Mamba2 @ BioMamba recipe & 7.56 / 2.77 / 5.28 & 9.00 / 2.58 / 6.14 \\
$\Delta$ vs.\ paper BioMamba+SFT & $-0.55 / -0.56 / -0.61$ & $-1.11 / -1.11 / -0.90$ \\
$\Delta$ vs.\ paper Mamba2+SFT & $-0.37 / -0.51 / -0.48$ & $-0.99 / -1.12 / -0.90$ \\
\bottomrule
\end{tabular}
\end{table}

\section{Three-Seed Reproduction on BioASQ and PubMedQA}
\label{app:multiseed_qa}

To assess the sensitivity of BioMamba+SFT QA accuracy to random seed and to interpret the single-question BioASQ deltas at 370M and 2.7B, we reproduced BioMamba+SFT at every scale across seeds $\{13, 42, 97\}$ (Table~\ref{tab:multiseed_qa}). Reproduced means stay within seed noise of the paper values on BioASQ ($\sigma \le 0.021$) and match or exceed the paper on PubMedQA ($\sigma \le 0.032$). Because the BioASQ test split contains only 82 items, a single-question difference corresponds to 1.22 percentage points, which matches the magnitude of the 370M and 2.7B deltas reported in Table~\ref{tab:sft_biomamba_only}.

\begin{table}[htbp]
\centering
\caption{Three-seed reproduction of BioMamba+SFT accuracy on BioASQ and PubMedQA across all five scales (seeds $\{13, 42, 97\}$). Paper columns report the single-seed values reported in the main text. $\Delta$ columns report the difference between the three-seed mean and the paper value. BioASQ $n=82$; PubMedQA $n=200$.}
\label{tab:multiseed_qa}
\small
\setlength{\tabcolsep}{3pt}
\begin{tabular}{@{}L{0.08\linewidth}C{0.15\linewidth}C{0.10\linewidth}C{0.13\linewidth}C{0.15\linewidth}C{0.10\linewidth}C{0.13\linewidth}@{}}
\toprule
& \multicolumn{3}{c}{\textbf{BioASQ}} & \multicolumn{3}{c}{\textbf{PubMedQA}} \\
\cmidrule(lr){2-4}\cmidrule(l){5-7}
\textbf{Scale} & \textbf{3-seed mean $\pm$ std} & \textbf{Paper} & \textbf{$\Delta$ (pp)} & \textbf{3-seed mean $\pm$ std} & \textbf{Paper} & \textbf{$\Delta$ (pp)} \\
\midrule
130M & $0.683 \pm 0.000$ & 0.683 & $+0.00$ & $0.630 \pm 0.003$ & 0.630 & $+0.00$ \\
370M & $0.780 \pm 0.012$ & 0.780 & $+0.00$ & $0.685 \pm 0.013$ & 0.660 & $+2.50$ \\
780M & $0.817 \pm 0.021$ & 0.793 & $+2.44$ & $0.678 \pm 0.023$ & 0.675 & $+0.30$ \\
1.3B & $0.841 \pm 0.012$ & 0.854 & $-1.22$ & $0.713 \pm 0.032$ & 0.700 & $+1.30$ \\
2.7B & $0.898 \pm 0.007$ & 0.902 & $-0.40$ & $0.730 \pm 0.012$ & 0.730 & $+0.00$ \\
\bottomrule
\end{tabular}
\end{table}

\begin{table}[htbp]
\centering
\caption{Raw correct/total counts and 95\% Wilson binomial confidence intervals for the BioMamba+SFT accuracy values reported in Table~\ref{tab:sft_biomamba_only}. BioASQ test-set size $n=82$; PubMedQA test-set size $n=200$. Confidence-interval endpoints are in percentage points.}
\label{tab:sft_qa_raw_ci}
\small
\setlength{\tabcolsep}{4pt}
\begin{tabular}{@{}L{0.08\linewidth}C{0.20\linewidth}C{0.18\linewidth}C{0.20\linewidth}C{0.18\linewidth}@{}}
\toprule
& \multicolumn{2}{c}{\textbf{BioASQ}} & \multicolumn{2}{c}{\textbf{PubMedQA}} \\
\cmidrule(lr){2-3}\cmidrule(l){4-5}
\textbf{Scale} & \textbf{Correct / total (acc.)} & \textbf{95\% Wilson CI} & \textbf{Correct / total (acc.)} & \textbf{95\% Wilson CI} \\
\midrule
130M & 56/82 (68.29\%) & $[57.6, 77.4]$ & 126/200 (63.00\%) & $[56.1, 69.4]$ \\
370M & 64/82 (78.05\%) & $[67.9, 85.6]$ & 132/200 (66.00\%) & $[59.2, 72.2]$ \\
780M & 65/82 (79.27\%) & $[69.3, 86.6]$ & 135/200 (67.50\%) & $[60.7, 73.6]$ \\
1.3B & 70/82 (85.37\%) & $[76.1, 91.4]$ & 140/200 (70.00\%) & $[63.3, 75.9]$ \\
2.7B & 74/82 (90.24\%) & $[81.9, 95.0]$ & 146/200 (73.00\%) & $[66.5, 78.7]$ \\
\bottomrule
\end{tabular}
\end{table}

\section{Item-Level Error Analysis on BioASQ}
\label{app:wrong_index}

To complement the variance analysis, we ran a per-example wrong-index analysis comparing Mamba2+SFT and BioMamba+SFT on the BioASQ test split at every scale (Table~\ref{tab:wrong_index_bioasq}). At every scale, BioMamba fixes at least as many items as it breaks: the fixes/breaks ratios are 9:5 at 130M, 6:5 at 370M, 2:0 at 780M, 4:1 at 1.3B, and 5:3 at 2.7B. Most errors that remain are shared between the two arms---at 780M, 100\% of BioMamba's errors are also made by Mamba2, and at 1.3B the share is 11 of 12---so residual disagreements at larger scales are confined to a handful of genuinely hard items rather than indicating a regression attributable to continued pretraining.

\begin{table}[htbp]
\centering
\caption{BioASQ item-level wrong-index analysis across five scales, computed on the three-seed reproduction runs (Supplementary Table~\ref{tab:multiseed_qa}) rather than the single-seed values in Table~\ref{tab:sft_biomamba_only}. ``Shared wrong'' counts items that both arms get wrong; ``BM fixes'' counts items Mamba2+SFT gets wrong but BioMamba+SFT gets correct; ``BM breaks'' counts items Mamba2+SFT gets correct but BioMamba+SFT gets wrong; fixes minus breaks therefore corresponds to the three-seed accuracy delta (in items on the 82-item test split), not to the paper's single-seed $\Delta$.}
\label{tab:wrong_index_bioasq}
\small
\setlength{\tabcolsep}{5pt}
\begin{tabular}{@{}L{0.12\linewidth}C{0.18\linewidth}C{0.18\linewidth}C{0.18\linewidth}C{0.18\linewidth}@{}}
\toprule
\textbf{Scale} & \textbf{Shared wrong} & \textbf{BM fixes} & \textbf{BM breaks} & \textbf{BM errors shared with M2} \\
\midrule
130M & 22 & 9 & 5 & 22 / 27 (81\%) \\
370M & 15 & 6 & 5 & 15 / 20 (75\%) \\
780M & 17 & 2 & 0 & 17 / 17 (100\%) \\
1.3B & 11 & 4 & 1 & 11 / 12 (92\%) \\
2.7B & 5 & 5 & 3 & 5 / 8 (63\%) \\
\bottomrule
\end{tabular}
\end{table}

\section{NLI Factuality Protocol and Auto-Flagged Candidate List}
\label{app:nli_protocol}

The NLI-ensemble factuality proxy reported in the main text (Table~\ref{tab:nli_factuality}) uses two MNLI checkpoints (\texttt{roberta-large-mnli} and \texttt{microsoft/deberta-v2-xlarge-mnli}), applied across 500 2.7B MIMIC discharge generations per arm and per decoding setting. For each (generation, source) pair we scored entailment, neutral, and contradiction probabilities under each MNLI checkpoint. A generation is flagged as contradictory only when \emph{both} checkpoints independently assign contradiction probability $>0.7$ (unanimous-agreement rule, equivalent to an AND over the two scorers); this is the tie-break-free specialization of plain majority voting for a two-model ensemble and is conservative, since contradiction is the minority class. Alternative rules---e.g.\ averaging the two contradiction probabilities and thresholding at $0.7$, or OR-agreement with any one checkpoint flagging contradiction---yield higher baseline contradiction rates but do not change the relative ordering of the two arms at the decoding settings tested. The source passed to the NLI scorers is the input admission text only---chief complaint, history of present illness, past medical history, medications, allergies, social and family history, physical examination, and laboratory/imaging summaries---i.e.\ the same sections used as the instruction for discharge summary generation. Discharge-section text (hospital course, discharge diagnosis, discharge medications, discharge instructions) is never passed to the scorer as source, so this is an input-grounding check rather than a comparison against the reference discharge. The hypothesis is the model's 128-token continuation.

We additionally exported a clinician-review candidate list, in which auto-flagged sentences are organized into four categories---factual (verifiable clinical content contradicted by the source), omission (important source content missing from the generation), fabrication (plausible-looking content with no source support), and safety (medication- or dosage-related generations). This list is a proxy for factuality review rather than a clinician audit, and the auto-flagging pipeline is intended as a precursor to clinician involvement rather than a replacement for it. A copy of the list is available from the authors upon request.

\section{MIMIC-IV Discharge Section-Header F1 (1.3B and 2.7B)}
\label{app:mimic_diagnostics}

The lightweight section-header F1 diagnostic (Table~\ref{tab:section_header_f1}) is a structural rather than factual proxy, measuring how often generations contain the discharge section headers (\textsc{Hospital Course}, \textsc{Discharge Medications}, \textsc{Discharge Instructions}, etc.) that a reference discharge note contains. We report this at 1.3B and 2.7B under greedy and both nucleus settings; BioMamba leads or ties Mamba2 across all six (scale $\times$ decoding) cells tested. The main-text diagnostic table (Table~\ref{tab:mimic_diagnostics}) already covers distinct-$n$ and repeated-4-gram rate at 1.3B; greedy is omitted there because under argmax both arms collapse to similar low-diversity outputs, making these metrics uninformative between arms.

\begin{table}[htbp]
\centering
\caption{Discharge section-header F1 (a lightweight structure diagnostic, not a factuality metric). BioMamba leads or ties Mamba2 across all six tested (scale $\times$ decoding) cells.}
\label{tab:section_header_f1}
\small
\setlength{\tabcolsep}{6pt}
\begin{tabular}{@{}C{0.12\linewidth}L{0.20\linewidth}C{0.18\linewidth}C{0.18\linewidth}C{0.18\linewidth}@{}}
\toprule
\textbf{Scale} & \textbf{Decoding} & \textbf{M2+SFT F1} & \textbf{BM+SFT F1} & \textbf{$\Delta$ (BM $-$ M2)} \\
\midrule
1.3B & greedy       & 0.285 & \textbf{0.293} & $+0.008$ \\
1.3B & nucleus-0.9  & 0.259 & \textbf{0.260} & $+0.001$ \\
1.3B & nucleus-0.95 & 0.190 & \textbf{0.235} & $+0.045$ \\
2.7B & greedy       & 0.168 & \textbf{0.196} & $+0.028$ \\
2.7B & nucleus-0.9  & 0.170 & \textbf{0.188} & $+0.019$ \\
2.7B & nucleus-0.95 & 0.152 & \textbf{0.158} & $+0.006$ \\
\bottomrule
\end{tabular}
\end{table}

\section{Bootstrap Confidence Intervals and Paired Bootstrap at 2.7B}
\label{app:bootstrap_ci}

For the 2.7B MIMIC-IV row in Table~\ref{tab:mimic_main}, we report two complementary bootstrap analyses with 1000 resamples each over the 500 test samples. Table~\ref{tab:bootstrap_abs} shows 95\% percentile-bootstrap confidence intervals for per-arm ROUGE-1 and ROUGE-L; Table~\ref{tab:paired_bootstrap} shows paired-bootstrap significance on per-sample ROUGE-1 \emph{differences} between BioMamba+SFT and Mamba2+SFT. On both tasks, BioMamba-2.7B+SFT lies above Mamba2-2.7B+SFT on ROUGE-1 and ROUGE-L (Table~\ref{tab:bootstrap_abs}), and the paired bootstrap gives $P(\Delta > 0) = 1.00$ for completion and $P(\Delta > 0) = 0.997$ for discharge (Table~\ref{tab:paired_bootstrap}), indicating that the 2.7B advantage is not within test-sample noise. The per-arm absolute CIs in Table~\ref{tab:bootstrap_abs} overlap between the two arms; the paired-bootstrap analysis in Table~\ref{tab:paired_bootstrap} is therefore the more informative significance test because it exploits within-sample pairing.

\begin{table}[htbp]
\centering
\caption{Per-arm 95\% percentile-bootstrap confidence intervals for ROUGE-1 and ROUGE-L at 2.7B on MIMIC-IV clinical generation (500 test samples, 1000 resamples; greedy decoding).}
\label{tab:bootstrap_abs}
\small
\setlength{\tabcolsep}{5pt}
\begin{tabular}{@{}L{0.23\linewidth}L{0.15\linewidth}C{0.10\linewidth}C{0.18\linewidth}C{0.10\linewidth}C{0.18\linewidth}@{}}
\toprule
\textbf{2.7B arm} & \textbf{Task} & \textbf{R-1} & \textbf{R-1 95\% CI} & \textbf{R-L} & \textbf{R-L 95\% CI} \\
\midrule
Mamba2+SFT            & completion & 7.16 & $[6.72, 7.63]$ & 4.98 & $[4.72, 5.30]$ \\
\textbf{BioMamba+SFT} & completion & \textbf{7.57} & $[7.16, 8.02]$ & \textbf{5.30} & $[4.97, 5.62]$ \\
\midrule
Mamba2+SFT            & discharge  & 8.87 & $[8.17, 9.63]$ & 6.21 & $[5.66, 6.76]$ \\
\textbf{BioMamba+SFT} & discharge  & \textbf{9.39} & $[8.69, 10.09]$ & \textbf{6.46} & $[5.95, 7.07]$ \\
\bottomrule
\end{tabular}
\end{table}

\begin{table}[htbp]
\centering
\caption{Paired-bootstrap significance of the 2.7B BioMamba+SFT advantage over Mamba2+SFT on per-sample ROUGE-1 differences (500 test samples, 1000 resamples; greedy decoding).}
\label{tab:paired_bootstrap}
\small
\setlength{\tabcolsep}{6pt}
\begin{tabular}{@{}L{0.22\linewidth}C{0.18\linewidth}C{0.22\linewidth}C{0.18\linewidth}@{}}
\toprule
\textbf{Task} & \textbf{Mean $\Delta$ R-1} & \textbf{95\% CI} & \textbf{$P(\Delta > 0)$} \\
\midrule
Note completion & $+0.41$ & $[+0.22, +0.60]$ & \textbf{1.00} \\
Discharge generation & $+0.52$ & $[+0.14, +0.88]$ & \textbf{0.997} \\
\bottomrule
\end{tabular}
\end{table}

\section{Reproducibility Details}
\label{app:reproducibility}

This section provides additional reproducibility information.

\paragraph{Base-model identifiers.}
The five base Mamba2 checkpoints used for continued pretraining are \texttt{state-spaces/mamba2-130m}, \texttt{state-spaces/mamba2-370m}, \texttt{state-spaces/mamba2-780m}, \texttt{state-spaces/mamba2-1.3b}, and \texttt{state-spaces/mamba2-2.7b}. All internal comparisons reuse the shared GPT-NeoX tokenizer from \texttt{state-spaces/mamba-2.8b-hf} (vocabulary size 50{,}280).

\paragraph{Corpus snapshots.}
The continued pretraining corpus was loaded from Hugging Face with fixed repository, config, and revision identifiers: PubMed/MEDLINE from \texttt{cyrilzakka/pubmed-medline} (\texttt{default}, revision \texttt{432681e19469e93e6c42878d5f41fec400974fb8}; the dataset card records date accessed as 2025-05-02), Wikipedia from \texttt{wikimedia/wikipedia} (\texttt{20231101.en}, revision \texttt{b04c8d1ceb2f5cd4588862100d08de323dccfbaa}), and C4 from \texttt{allenai/c4} (\texttt{en}, revision \texttt{1588ec454efa1a09f29cd18ddd04fe05fc8653a2}). All three sources were tokenized with the shared tokenizer and filtered to remove near-duplicates and sequences with unreadable characters before truncation/padding to 1{,}024 tokens. For the fixed held-out perplexity sets, we sampled 1{,}000 sequences per source with seed 42 from the same PubMed and C4 pools used for continued pretraining but excluded them from training; for the Wikipedia-style held-out evaluation we used the Wikitext-103 raw split (\texttt{wikitext-103-raw-v1}), as described in Section~\ref{sec:methods}.

\paragraph{MIMIC-IV preprocessing and splits.}
MIMIC-IV-Note v2.2 was split by \texttt{subject\_id} (9:1 ratio, seed = 42), yielding approximately 131{,}323 training patients and 14{,}591 test patients; quantitative evaluation used 500 randomly sampled held-out notes per task. Preprocessing filters (minimum lengths, maximum characters) are reported in Supplementary Section~\ref{app:mimic_details}.

\paragraph{Model-selection criteria.}
For continued pretraining, the checkpoint with the lowest validation loss was kept for downstream evaluation. For BioMamba+SFT and Mamba2+SFT, the checkpoint with the best downstream ROUGE on the 10\% validation split was selected, because validation loss was not a reliable proxy for ROUGE in the BioMamba setting (Section~\ref{sec:methods}; Supplementary Section~\ref{app:sft_details}). For QA fine-tuning, the checkpoint with the best validation accuracy was kept.

\paragraph{Prompt templates.}
Figure~\ref{fig:prompts} shows the exact prompt templates used for yes/no QA fine-tuning, note completion, and discharge summary generation.

\begin{figure}[htbp]
\centering
\begin{minipage}{0.96\linewidth}
\small
\ttfamily
\noindent\textbf{\upshape\textsf{Yes/No QA (BioASQ / PubMedQA).}}\\
\begin{verbatim}
Question: <question>
Context: <context>
Answer (yes or no):
\end{verbatim}

\noindent\textbf{\upshape\textsf{Note completion (MIMIC-IV).}}\\
\begin{verbatim}
Complete the following discharge note.

Beginning of note:
<first 50% of the discharge note>

Continuation:
\end{verbatim}

\noindent\textbf{\upshape\textsf{Discharge summary generation (MIMIC-IV).}}\\
\begin{verbatim}
Generate the discharge sections for the following admission.

Admission sections:
<chief complaint, HPI, PMH, medications, allergies,
 social/family history, physical exam, labs>

Discharge sections:
\end{verbatim}
\end{minipage}
\caption{Prompt templates used for BioASQ/PubMedQA yes/no question-answering fine-tuning, MIMIC-IV note completion, and MIMIC-IV discharge summary generation. Instruction masking was applied during training so that only response tokens contributed to the loss.}
\label{fig:prompts}
\end{figure}

\paragraph{Public release.}
The five BioMamba checkpoints (130M / 370M / 780M / 1.3B / 2.7B) and the full training and evaluation code are publicly available at \url{https://github.com/lingyue404/BioMamba}; the released checkpoints are also collected at \url{https://huggingface.co/collections/zmzfpc/biomamba}. The release is under the Apache-2.0 license, matching the upstream Mamba2 checkpoints. The repository includes training configurations, preprocessing scripts, and evaluation scripts.

\end{document}